\documentclass[10pt,twocolumn,letterpaper]{article}

\usepackage{iccv}
\usepackage{times}
\usepackage{epsfig}
\usepackage{graphicx}
\usepackage{amsmath}
\usepackage{amssymb}
\usepackage{booktabs}
\usepackage{makecell}
\usepackage{arydshln}
\usepackage{enumitem}
\usepackage[table]{xcolor}
\usepackage{pifont}
\newcommand{\cmark}{\ding{51}}%
\usepackage{multirow}
\usepackage{subcaption}

\newcommand{\myparagraph}[1]{\vspace{3pt}\noindent\textbf{#1}}
\newcommand\our{\text{Mask2Anomaly}}

\usepackage[pagebackref=true,breaklinks=true,letterpaper=true,colorlinks,bookmarks=false]{hyperref}
\usepackage[capitalize]{cleveref}
\crefname{section}{Sec.}{Secs.}
\Crefname{section}{Section}{Sections}
\Crefname{table}{Table}{Tables}
\crefname{table}{Tab.}{Tabs.}
\iccvfinalcopy 

\ificcvfinal\fi

\begin{document}

\title{Unmasking Anomalies in Road-Scene Segmentation}

\author{Shyam Nandan Rai$^{1}$, Fabio Cermelli$^{1,2}$, Dario Fontanel$^{1}$, Carlo Masone$^{1}$, Barbara Caputo$^{1}$\\
$^{1}$Politecnico di Torino, $^{2}$Italian Institute of Technology\\
{\tt\small {first.last}@polito.it}
}
\maketitle
\ificcvfinal\thispagestyle{empty}\fi

\begin{abstract}
Anomaly segmentation is a critical task for driving applications, and it is approached traditionally as a per-pixel classification problem. However, reasoning individually about each pixel without considering their contextual semantics results in high uncertainty around the objects' boundaries and numerous false positives. We propose a paradigm change by shifting from a per-pixel classification to a mask classification. Our mask-based method, {\our}, demonstrates the feasibility of integrating an anomaly detection method in a mask-classification architecture. {\our} includes several technical novelties that are designed to improve the detection of anomalies in masks: i) a global masked attention module to focus individually on the foreground and background regions; ii) a mask contrastive learning that maximizes the margin between an anomaly and known classes; and iii) a mask refinement solution to reduce false positives. Mask2Anomaly achieves new state-of-the-art results across a range of benchmarks, both in the per-pixel and component-level evaluations. In particular, {\our} reduces the average false positives rate by 60\% \wrt the previous state-of-the-art. Github page: \href{https://github.com/shyam671/Mask2Anomaly-Unmasking-Anomalies-in-Road-Scene-Segmentation}{https://tinyurl.com/54ydrxvj}
\end{abstract}


\section{Introduction}
\label{sec:intro}
Semantic segmentation~\cite{Cordts2016Cityscapes,sun2019naae, Zhang2019fsda,yang2020fda,tavera2022pixda} plays a significant role in self-driving cars because it provides a detailed understanding of surroundings. 
Generally, semantic segmentation models are trained to recognize a pre-defined set of semantic classes (\eg car, pedestrian, road, etc.); however, in real-world applications, they may encounter objects not belonging to such categories (\eg animals or cargo dropped on the road). Therefore, it is essential for these models to identify objects in a scene that are not present during training \ie \textit{anomalies}, both to avoid potential dangers and to enable continual learning~\cite{michieli2021continual, cermelli2020modelingthebackground, douillard2020plop, cermelli2022incremental} and open-world solutions~\cite{cen2021deep}.

\begin{figure}[t]
    \begin{center}
        \includegraphics[width=1\linewidth]{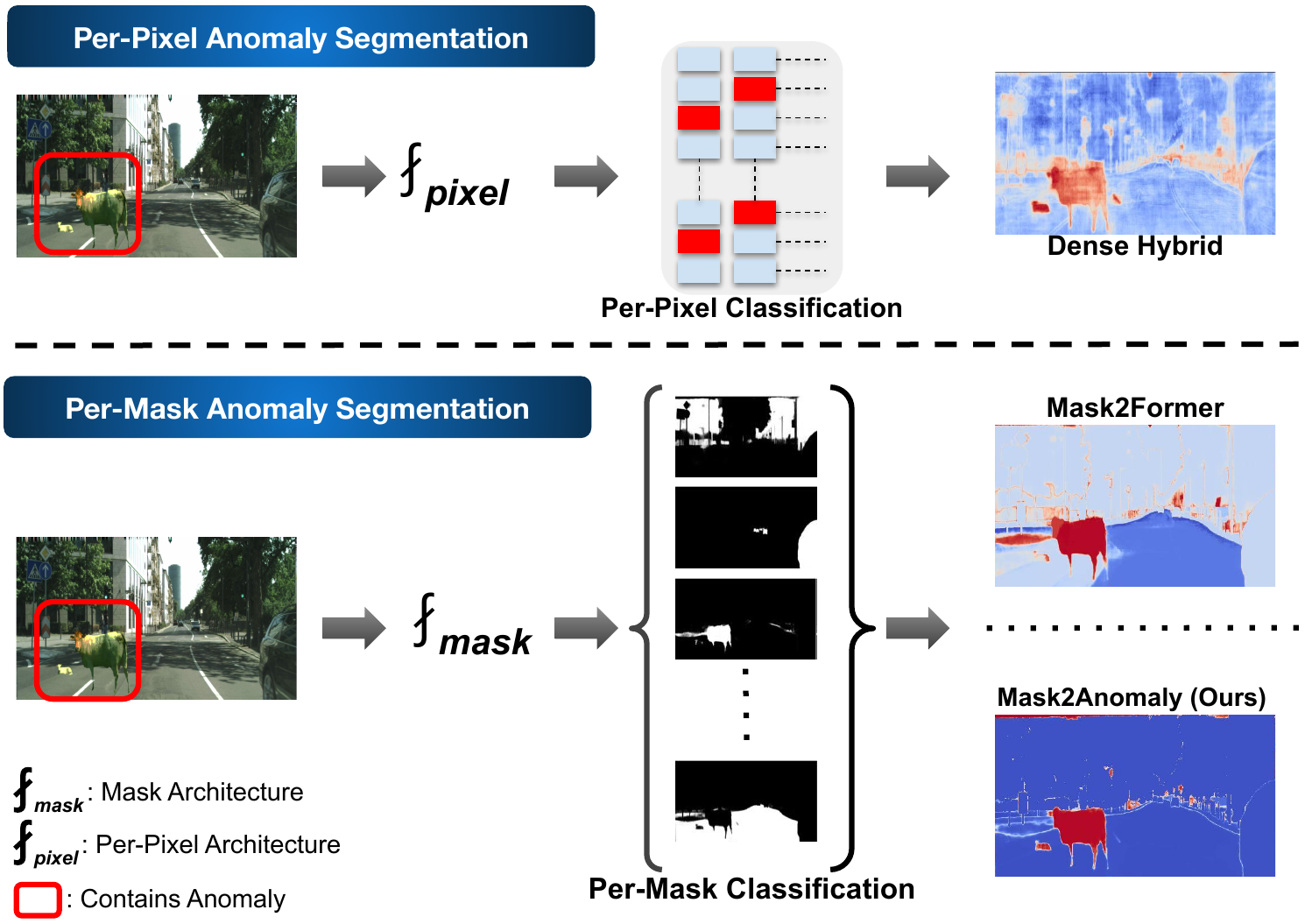}
    \end{center}
    \vspace{-1em}
      \caption{\textbf{Per-pixel vs per-mask Anomaly Segmentation:} Dense Hybrid~\cite{grcic2022densehybrid}, the state-of-the-art method for AS based on per-pixel classification can detect the anomalies, but it produces many false positives. Anomaly segmentation can be cast as a mask classification problem, but naively using MSP~\cite{hendrycks2016baseline} on top of Mask2Former~\cite{cheng2022masked} does not produce good results. Our Mask2Anomaly exploits mask-transformers properties to refine the classification of anomalies, drastically reducing false positives. $f_{pixel}$ and $f_{mask}$ denotes per-pixel, and per-mask architecture. Anomalies in the output image are represented in {red}.}
    \label{fig:teaser} 
\end{figure}

Anomaly segmentation (AS)~\cite{blum2019fishyscapes,xia2020synth,fontanel2021detecting,jung2021standardized} addresses this problem, \ie it aims to segment objects from classes that were absent during training. Existing AS methods are built upon the idea of individually classifying the pixels and assigning to each of them an anomaly score. This score may be given by a pixel-level discriminative method~\cite{angus2019oodsem, jung2021standardized, grcic2022densehybrid, tian2022pixel}, by estimating the uncertainty of the individual pixel predictions~\cite{mukhoti2018evaluating}, or by comparing the per-pixel discrepancy between the original image and a synthetic image generated from the semantic predictions~\cite{lis2019detecting, vojir2021road, xia2020synthesize}.
However, reasoning on the pixels individually produces noisy anomaly scores, thus leading to a high number of false positives and poorly localized anomalies (see \cref{fig:teaser}).

In this paper, we propose to address this problem by casting AS  as a mask classification task rather than a pixel classification. This idea stems from the recent advances in mask-transformer architectures~\cite{cheng2022masked, cheng2021per}, which demonstrated that it is possible to achieve remarkable performance across various segmentation tasks by classifying masks, rather than pixels. We hypothesize that mask-transformer architectures are better suited to detect anomalies than per-pixel architectures~\cite{chen2018encoder,huang2017densely}, because masks encourage objectness and thus can capture anomalies as whole entities, leading to more congruent anomaly scores and reduced false positives. 
To enable the segmentation of anomalies at the mask level, we revisit the Maximum Softmax Probability (MSP)~\cite{hendrycks2016baseline}, a classic method used in per-pixel AS, and apply it to the masks produced by a mask-transformer model.
However, the effectiveness of such an approach hinges on the model's capability to output masks that capture well anomalies and we found that naively using MSP on top of the best mask-transformer architecture~\cite{cheng2022masked} does not yield good results (see \cref{fig:teaser}).
Hence, we propose several technical contributions to improve the capability of mask-transformer architectures to capture anomalies and reject false positives in driving scenes (see \cref{fig:teaser}): 
\begin{itemize}[noitemsep,topsep=0pt]
    \item At the \textbf{architectural} level, we propose a global masked-attention mechanism that allows the model to focus on both the foreground objects and on the background while retaining the efficiency of the original masked-attention~\cite{cheng2022masked}.
    \item At the \textbf{training} level, we have developed a mask contrastive learning framework that utilizes outlier masks from additional out-of-distribution data to maximize the separation between anomalies and known classes.
    \item At the \textbf{inference} level, we propose a mask-based refinement solution that reduces false positives by filtering masks based on the panoptic segmentation~\cite{kirillov2019panoptic} that distinguishes between ``things'' and ``stuff''.
\end{itemize}
We integrate these contributions on top of the mask architecture~\cite{cheng2022masked} and term this solution \textbf{\our}.
To the best of our knowledge, {\our} is the first demonstration of an AS method that detects anomalies at the mask level. We tested {\our} on standard anomaly segmentation benchmarks for road scenes (Road Anomaly~\cite{lis2019detecting}, Fishyscapes~\cite{blum2021fishyscapes},  Segment Me If You Can~\cite{chan2021segmentmeifyoucan}), achieving the best results among all AS methods by a significant margin. In particular, {\our} reduces on average the false positives rate by more than half \wrt the previous state-of-the-art. Code and pre-trained models will be made publicly available upon acceptance.

\section{Related Work}
\label{sec:related}
\myparagraph{Mask-based semantic segmentation.}
Traditionally, semantic segmentation methods \cite{long2015fully, chen2018encoder, zhao2017pyramid, lin2017refinenet, zhang2018exfuse} have adopted fully-convolutional encoder-decoder architectures \cite{long2015fully, badrinarayanan2017segnet} and addressed the task as a dense classification problem. 
However, transformer architectures have recently caused us to question this paradigm due to their outstanding performance in closely related tasks such as object detection~\cite{carion2020end} and instance segmentation~\cite{he2017mask}. In particular, \cite{cheng2021per} proposed a mask-transformer architecture that addresses segmentation as a mask classification problem. It adopts a transformer and a per-pixel decoder on top of the feature extraction. The generated per-pixel and mask embeddings are combined to produce the segmentation output. Building upon~\cite{cheng2021per},~\cite{cheng2022masked} introduced a new transformer decoder adopting a novel masked-attention module and feeding the transformer decoder with one pixel-decoder high-resolution feature at a time. 

So far, all these mask-transformers have been considered exclusively in a closed set setting, i.e, there are no unknown categories at test time. 
To the best of our knowledge, Mask2Anomaly is the first method that performs AS directly with mask-transformers, thus empowering these approaches with the capability to recognize anomalies in real-world settings.

\begin{figure*}[t]
\begin{center}

\includegraphics[width=1\linewidth, height=0.45\linewidth]{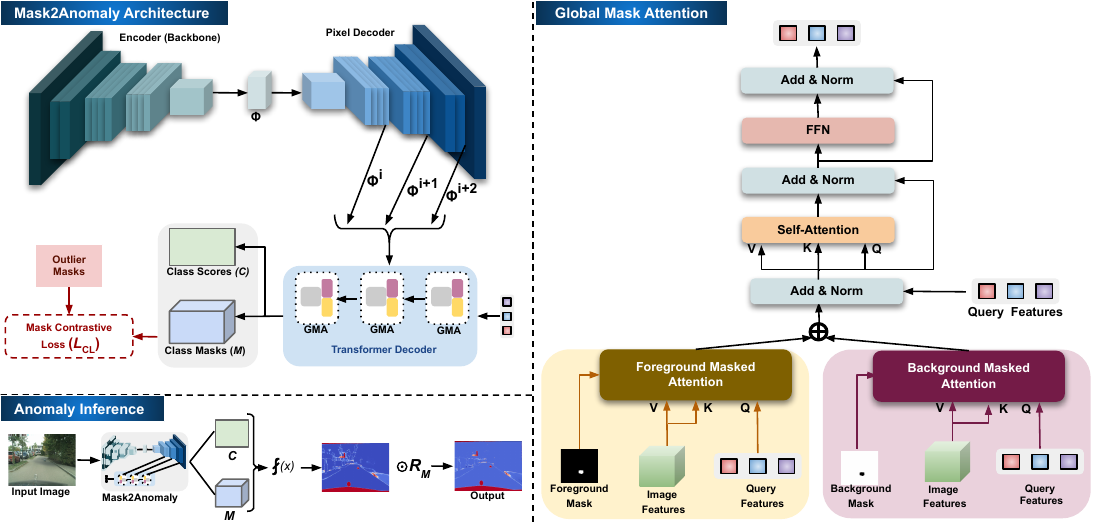}
\end{center}
\vspace{-1em}
  \caption{\textbf{Mask2Anomaly Overview.} Mask2Anomaly meta-architecture consists of an encoder, a pixel decoder, and a transformer decoder. We propose global mask attention (\cref{sec:global_attention}) that independently distributes the attention between foreground and background. V, K, and Q are Value, Key, and Query. $\phi$ is image features. $\phi^{i}, \phi^{i+1}, \phi^{i+2}$ are upsampled image features at multiple scales. Mask contrastive Loss $L_{CL}$ (\cref{subsec:CL}) utilizes outlier masks to maximize the separation between anomalies and known classes. During anomaly inference, we utilize refinement mask $R_{M}$ (\cref{subsec:RM}) to minimize false positives.} \vspace{-1em}
\label{fig:overview}
\end{figure*}

\myparagraph{Anomaly segmentation} methods can be broadly divided into three categories: (a) Discriminative, (b) Generative, and (c) Uncertainty-based methods.~\textit{Discriminative Methods} are based on the classification of the model outputs. Hendrycks and Gimpel~\cite{hendrycks2016baseline} established the initial AS discriminative baseline by applying a threshold over the maximum softmax probability (MSP) that distinguishes between in-distribution and out-of-distribution data. Other approaches use auxiliary datasets to improve performance \cite{liang2017enhancing, jung2021standardized,tian2022pixel} by calibrating the model over-confident outputs.
Alternatively, \cite{lee2018simple} learns a confidence score by using the Mahalanobis distance, and \cite{chan2021entropy} introduces an entropy-based classifier to discover out-of-distribution classes.
Recently, discriminative methods tailored for semantic segmentation \cite{blum2021fishyscapes} directly segment anomalies in embedding space. In contrast, \cite{grcic2022densehybrid} proposes a hybrid approach that combines the known class posterior, dataset posterior, and an un-normalized data likelihood to estimate anomalies. 
\textit{Generative Methods } provides an alternative paradigm to segment anomalies based on generative models ~\cite{lis2019detecting, di2021pixel, xia2020synthesize, vojir2021road}. These approaches train generative networks to reconstruct anomaly-free training data and then use the generation discrepancy to detect an anomaly at test time. All the generative-based methods heavily rely on the generation quality and thus experience performance degradation due to image artifacts~\cite{fontanel2021detecting}.
Finally,~\textit{Uncertainty based} methods segment anomalies by leveraging uncertainty estimates via Bayesian neural networks~\cite{mukhoti2018evaluating}. 

All the methods discussed above are based on per-pixel classification architectures and score the pixels individually without considering local semantics, leading to noisy anomaly predictions and many false positives. {\our} overcomes this limitation by segmenting anomalies as semantically clustered masks, encouraging the objectness of the predictions. To the best of our knowledge, this is the first work to use masks to score anomalies.

\section{Method}
\label{sec:methodology}
In this section, we begin by introducing problem-setting, followed by describing a generic mask-transformer architecture for anomaly segmentation. Next, we delve into our {\our} architecture and its novel elements.

\subsection{Preliminaries}
Let us denote with $\mathcal{X} \subset \mathbb{R}^{3 \times H \times W}$ the space of RGB images, where $H$ and $W$ are the height and width, respectively, and with $\mathcal{Y} \subset \mathbb{N}^{K \times H \times W}$ the space of semantic labels that associate each pixel in an image to a semantic category from a predefined set $\mathcal{K}$, with $|\mathcal{K}|=K$. At training time we assume to have a dataset $\mathcal{D} = \left\{ (x_i, y_i)\right\}_{i=1}^{D}$, where  $x_i \in \mathcal{X}$ is an image and $y_i\in \mathcal{Y}$ is its ground truth semantic mask. 
The goal for an anomaly segmentation model is to learn a function $f$ that maps the image space to an anomaly score space, \ie $f: \mathcal{X} \mapsto \mathbb{R}^{H \times W}$. For traditional semantic segmentation architectures based on per-pixel classification~\cite{chen2018encoder}, the function $f$ can be obtained in various ways, for example, applying the \textit{Maximum Softmax Probability} (MSP)~\cite{hendrycks2016baseline} on top of the per-pixel classifier. Formally, given the pixel-wise class scores $S(x) \in [0,1]^{K\times H \times W}$ obtained by segmenting the image $x$ with a per-pixel architecture, we compute the anomaly score as:
\begin{equation}
    \label{eq:perpixel_msp}
    f(x) = 1-\max_{k=1}^{K}(S(x)).
\end{equation}

In this paper, we propose to adapt this framework based on MSP to mask-transformer segmentation architectures. We recall that the mask classification problem is formulated as a direct set prediction task with the goal of producing a fixed-size set of $N$ predictions \cite{carion2020end}.
Based on this idea, the mask classification meta-architecture for semantic segmentation consists of three parts: a) a \textit{backbone} that acts as feature extractor, b) a \textit{pixel-decoder} that upsamples the low-resolution features extracted from the backbone to produce high-resolution \textit{per-pixel embeddings}, and c) a \textit{transformer decoder}, made of $L$ transformer layers, that takes the image features to output a fixed number of object queries consisting of \textit{mask embeddings} and their associated \textit{class scores} $C\in \mathbb{R}^{N \times K}$. The final \textit{class mask} $M\in \mathbb{R}^{N\times (H \times W)}$ are obtained by multiplying the mask embeddings with the per-pixel embeddings. 
The mask-transformer is trained using a combination of binary cross-entropy loss and dice loss~\cite{milletari2016v} for the class masks and cross-entropy loss for the class scores, unlike per-pixel architecture that is trained only on cross-entropy loss (more details on these losses are given in the supplementary material).

Given such a mask-transformer architecture, we propose to calculate the 
anomaly scores for an input $x$ as 
\begin{equation}
    \label{eq:mask_msp}
    f(x) = 1 - \max_{k=1}^{K} \left(\text{softmax}(C)^T \cdot \text{sigmoid}(M)\right).
\end{equation}
Here, $f(x)$ utilizes the same marginalization strategy of
class and mask pairs as~\cite{cheng2021per} to get anomaly scores. 
Without loss of generality, we implement the anomaly scoring (\cref{eq:mask_msp}) on top of the Mask2Former~\cite{cheng2022masked} architecture. However, this strategy hinges on the fact that the masks predicted by the segmentation architecture can capture anomalies well. We found that simply applying the MSP on top of Mask2Former as in \cref{eq:mask_msp} does not yield good results (see \cref{fig:teaser} and the results in \cref{sec:ablation}). To overcome this problem, we introduce improvements in the architecture, training procedure, and anomaly inference mechanism. We name our method as {\our},  and its overview is shown in \cref{fig:overview} (left). In the rest of the sections, we will discuss in detail the technical novelties of {\our}.

\begin{figure}[t]
    \begin{center}
        \includegraphics[width=1\linewidth]
        {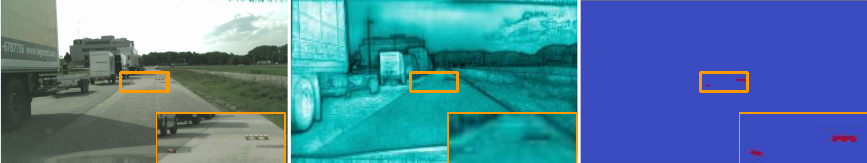}
    \end{center}
    \vspace{-1.5em}
    {\small{\hspace{2.25em}Input Image \hspace{2.75em}Attention Map\hspace{3.25em}Ground Truth}}
\caption{\textbf{Limitation of Mask-Attention:} Masked-attention~\cite{cheng2022masked}
selectively attends to foreground regions resulting in low attention scores (dark regions) for anomalies. Anomalies are in~{red}. Best viewed with zoom.}
    \label{fig:mask2former_attention}
    \vspace{-1em}
\end{figure}

\subsection{Global Masked Attention}
\label{sec:global_attention}
One of the key ingredients to Mask2Former~\cite{cheng2022masked} state-of-the-art segmentation results is the replacement of the \textit{cross-attention} (CA) layer in the transformer decoder with a \textit{masked-attention} (MA). The masked-attention attends only to pixels within the foreground region of the predicted mask for each query, under the hypothesis that local features are enough to update the query object features. 
The output of the $l$-th masked-attention layer can be formulated as
\begin{equation}
    \text{softmax}(\mathcal{M}^F_l + QK^{T})V + X_{in} 
\end{equation}
where $X_{in}\in$ $\mathbb{R}^{N \times C}$ are the $N$ $C$-dimensional query features from the previous decoder layer. The input queries $Q \in \mathbb{R}^{N \times C}$ are obtained by linearly transforming the query features with a learnable transformation whereas the keys and values $K, V$ are the image features under learnable linear transformations $f_k(.)$ and $f_v()$. 
Finally, $\mathcal{M}_l^F$ is the predicted foreground attention mask that at each pixel location $(i, j)$ is defined as
\begin{equation}
    \mathcal{M}_l^F(i, j) = \begin{cases}
        0 & \text{if } M_{l-1}(i, j)  \geq 0.5 \\
        -\infty & \text{otherwise},
    \end{cases}
\end{equation}
where $M_{l-1}$ is the output mask of the previous layer. 

By focusing only on the foreground objects, masked-attention grants faster convergence and better semantic segmentation performance than cross-attention. However, focusing only on the foreground region constitutes a problem for anomaly segmentation because anomalies may also appear in the background regions. Removing background information leads to failure cases in which the anomalies in the background are entirely missed, as shown in the example in \cref{fig:mask2former_attention}. To ameliorate the detection of anomalies in these corner cases, we extend the masked attention with an additional term focusing on the background region (see \cref{fig:overview}, right). We call this a \textit{global masked-attention} (GMA) formally expressed as 
\begin{equation}
    \label{eq:gma}
    \begin{aligned}
        X_{out} = &\text{softmax}(\mathcal{M}_l^{F} + QK^{T})V \\
        +& \text{softmax}(\mathcal{M}_l^{B} + QK^{T})V + X_{in}
    \end{aligned}
\end{equation}
where $ \mathcal{M}_l^B$ is the additional background attention mask that complements the foreground mask $\mathcal{M}_l^F$, and it is defined at the pixel coordinates $(i, j)$ as
\begin{equation}
    \mathcal{M}_l^B(i, j) = \begin{cases}
        0 & \text{if } M_{l-1}(i, j)  < 0.5 \\
        -\infty & \text{otherwise}.
    \end{cases}
\end{equation}

The global masked-attention in \cref{eq:gma} differs from the masked-attention by additionally attending to the background mask region, yet it retains the benefits of faster convergence w.r.t. the cross-attention.


\subsection{Mask Contrastive Learning}
\label{subsec:CL}
The ideal characteristic of an anomaly segmentation model is to predict high anomaly scores for out-of-distribution (OOD) objects and low anomaly scores for in-distribution (ID) regions. Namely, we would like to have a significant margin between the likelihood of known classes being predicted at anomalous regions and vice-versa.
A common strategy used to improve this separation is to fine-tune the model with auxiliary out-of-distribution (anomalous) data as supervision~\cite{grcic2020dense,grcic2022densehybrid, blum2021fishyscapes}.  

Here we propose a contrastive learning approach to encourage the model to have a significant margin between the anomaly scores for in-distribution and out-of-distribution classes. 
Our mask-based framework allows us to straightforwardly implement this contrastive strategy by using as supervision 
outlier images generated by cutting anomalous objects from the auxiliary OOD data and pasting it on top of the training data. For each outlier image, we can then generate a binary outlier mask $M_{OOD}$ that is $1$ for out-of-distribution pixels and $0$ for in-distribution class pixels. With this setting, we first calculate the negative likelihood of in-distribution classes using the class scores $C$ and class masks $M$ as:
\begin{equation}
    \begin{aligned}
         l_{N} = - \max_{k=1}^{K} \left(\text{softmax}(C)^T \cdot \text{sigmoid}(M)\right)
    \end{aligned}
\end{equation}
Ideally,  for pixels corresponding to in-distribution classes $l_{N}$ should be $-1$ since the value of $\text{softmax}(C)^T$ and $\text{sigmoid}(M)$ would be close to $1$. On the other hand, for an anomalous pixel, $\text{sigmoid}(M)$ is ideally 0 as M contains only inlier classes mask that results in $l_{N}$ to be $0$. Using $l_{N}$, we define our contrastive loss as:
\begin{equation}
    \begin{aligned}
         L_{CL} &= \frac{1}{2}(l_{CL}^{2}),\\
          l_{CL} &= 
    \begin{cases}
        l_{N} & \text{if} M_{OOD} = 0 \\
        max(0, m - l_{N}) & \text{otherwise,}
    \end{cases}
    \end{aligned}
\end{equation}
where the margin $m$ is a hyperparameter that decides the minimum distance between the out-of-distribution and in-distribution classes.
\begin{figure}[t]
    \begin{center}
        \includegraphics[width=1\linewidth]{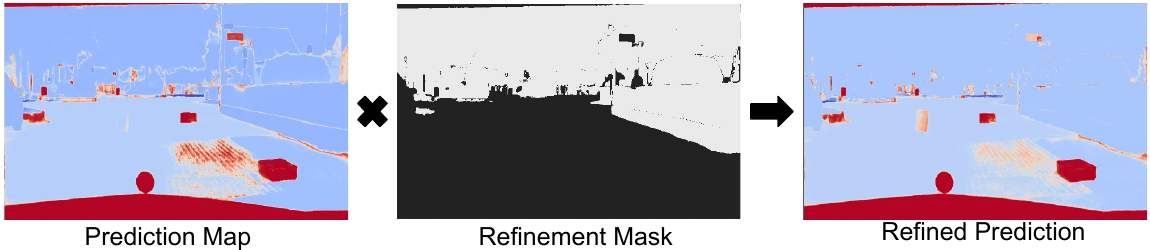}
    \end{center}
    \vspace{-1em}
    \caption{\textbf{Mask Refinement Illustration:} To obtain the refined prediction, we multiply the prediction map with a refinement mask that is built by assigning zero anomaly scores for pixels that are categorized as ``stuff'', except for the ``road''. The refinement eliminates many false positives at the boundary of objects and in the background. The region to be masked is white in the refinement mask.}
    \vspace{-1em}
    \label{fig:refinements}
\end{figure}
\subsection{Refinement Mask}
\label{subsec:RM}
False positives are one of the main problems in anomaly segmentation, particularly around object boundaries. Handcrafted methods such as iterative boundary suppression ~\cite{jung2021standardized} or dilated smoothing have been proposed to minimize the false positives at boundaries or globally, however, they require tuning for each specific dataset. 
Instead, we propose a general 
refinement technique that leverages the capability of mask transformers~\cite{cheng2022masked} to perform all segmentation tasks.
Our method stems from the panoptic perspective~\cite{kirillov2019panoptic} that the elements in the scene can be categorized as \textit{things}, \ie countable objects, and \textit{stuff}, \ie amorphous regions. 
With this distinction in mind, we observe that in driving scenes, i) unknown objects are classified as things, and ii) they are often present on the road. Thus, we can proceed to remove most false positives by filtering out all the masks corresponding to “stuff", except the ``road" category. We implement this removal mechanism in the form of a binary refinement mask $R_{M} \in [0,1]^{H \times W}$, which contains zeros in the segments corresponding to the unwanted ``stuff" masks and one otherwise. Thus, by multiplying $R_{M}$ with the predicted anomaly scores $f$ we filter out all the unwanted ``stuff" masks and eliminate a large portion of the false positives (see \cref{fig:refinements}). Formally, for an image $x$ the refined anomaly scores $f^r$ is computed as:
\begin{equation}
    f^r(x) = R_{M} \odot f(x),
\end{equation}
where $\odot$ is the Hadamard product. 

$R_M$ is the dot product between the binarized output mask $\bar{M} \in \{0,1\}^{N \times (H \times W)}$ and the class filter $\bar{C} \in \{0,1\}^{1 \times N}$, \ie $R_M = \bar{C} \cdot \bar{M}$.
We define $\bar{M} = \text{sigmoid}(M) > 0.5$ and the class filter $\bar{C}$ is equal to $1$ only where the highest class score of $\text{softmax}(C)$ belongs to ``things" or ``road" classes and is greater than $0.95$. 

\section{Experiments}
\myparagraph{Dataset:} We train Mask2Anomaly on Cityscapes~\cite{Cordts2016Cityscapes} and for evaluation we use Road Anomaly~\cite{lis2019detecting},
Fishyscapes~\cite{blum2019fishyscapes} and Segment Me If You Can (SMIYC) benchmarks~\cite{chan2021segmentmeifyoucan}. 

\noindent\textit{Road Anomaly:} is a collection of 60 web images having anomalous objects located on or near the road. 

\noindent\textit{Fishyscapes (FS):} consists of two datasets, Fishyscape static (FS static) and Fishyscapes lost \&  found (FS lost \&  found). Fishyscape static is built by blending Pascal VOC~\cite{everingham2010pascal} objects on Cityscapes images containing 30 validation and 1000 test images. Fishyscapes lost \& found is based on a subset of the Lost and Found dataset~\cite{pinggera2016lost}, with 100 validation and 275 test images. 

\noindent\textit{SMIYC:} consists of two datasets, RoadAnomaly21 (SMIYC-RA21) and RoadObstacle21 (SMIYC-RO21). The SMIYC-RA21  contains 10 validation and 100 test images with diverse anomalies. The SMIYC-RO21 is collected with a focus on segmenting road anomalies and has 30 validation and 327 test images.

\begin{table*}[!ht]
\centering
\renewcommand{\arraystretch}{1}
\resizebox{1\linewidth}{!}{\begin{tabular}{c|cc|cc|cc|cc|cc|cc}
\multicolumn{1}{c|}{}&\multicolumn{2}{c|}{SMIYC RA-21}&\multicolumn{2}{c|}{SMIYC RO-21}&\multicolumn{2}{c|}{FS L\&F}&\multicolumn{2}{c|}{FS Static}&\multicolumn{2}{c|}{Road Anomaly}& \multicolumn{2}{c}{Average}\\
\cline{2-3}\cline{4-5}\cline{6-7}\cline{8-9}\cline{10-11}\cline{12-13}
\multicolumn{1}{c|}{Methods}&\multicolumn{1}{c|}{AuPRC $\uparrow$}&\multicolumn{1}{c|}{FPR$_{95}$ $\downarrow$}&\multicolumn{1}{c|}{AuPRC $\uparrow$} &\multicolumn{1}{c|}{FPR$_{95}$ $\downarrow$}&\multicolumn{1}{c|}{AuPRC $\uparrow$} &\multicolumn{1}{c|}{FPR$_{95}$ $\downarrow$}&\multicolumn{1}{c|}{AuPRC $\uparrow$} &\multicolumn{1}{c|}{FPR$_{95}$ $\downarrow$}&\multicolumn{1}{c|}{AuPRC $\uparrow$} &\multicolumn{1}{c|}{FPR$_{95}$ $\downarrow$}&\multicolumn{1}{c|}{AuPRC $\uparrow$} &\multicolumn{1}{c}{FPR$_{95}$$\downarrow$}\\
\Xhline{3\arrayrulewidth}
Max Softmax~\cite{hendrycks2016baseline}(ICLR'17) &27.97	&72.02	&15.72	&16.6	&1.77	&44.85	&12.88	&39.83	&15.72	&71.38	&14.81	&48.93\\
Entropy~\cite{hendrycks2016baseline}(ICLR'17) &-	&-	&-	&-	&2.93	&44.83	&15.4	&39.75	&16.97	&71.1	&11.66&	51.89\\
Mahalanobis~\cite{lee2018simple}(NeurIPS'18) &20.04	&86.99	&20.9	&13.08	&-	&-	&-	&-	&14.37	&81.09	&18.42	&60.38\\

Image Resynthesis~\cite{lis2019detecting}(ICCV'19) &52.28	&25.93	&37.71	&4.7&	5.7	&48.05	&29.6	&27.13	&-	&-	&31.32	&26.45\\
Learning Embedding~\cite{blum2021fishyscapes}(IJCV'21) &37.52	&70.76	&0.82	&46.38	&4.65	&24.36	&57.16	&13.39	&-	&-	&26.18	&45.43\\

Void Classifier~\cite{blum2021fishyscapes}(IJCV'21)&36.61	&63.49	&10.44	&41.54	&10.29	&22.11	&4.5	&19.4	&-	&-	&15.46	&36.63\\

JSRNet~\cite{vojir2021road}(ICCV'21)&33.64	&43.85&	28.09	&28.86	&-	&-	&-	&-	&\textbf{94.4}	&\textbf{9.2}	&52.04	&47.3\\
SML~\cite{jung2021standardized}(ICCV'21) &46.8	&39.5	&3.4	&36.8	&31.67	&21.9	&52.05	&20.5	&17.52	&70.7	&30.28	&37.88\\
SynBoost~\cite{di2021pixel}(CVPR'21) &56.44	&61.86	&71.34	&3.15	&43.22	&15.79	&72.59	&18.75	&38.21	&64.75	&56.36	&32.86\\
Maximized Entropy~\cite{chan2021entropy}(ICCV'21) &\underline{85.47}	&15.00	&85.07	&0.75	&29.96	&35.14	&86.55	&8.55	&48.85	&31.77	&\underline{67.18}	&18.24\\
Dense Hybrid~\cite{grcic2022densehybrid}(ECCV'22)&77.96&	\textbf{9.81}	&\underline{87.08}	&\underline{0.24}	&\textbf{47.06}	&\textbf{3.97}&	80.23	&5.95	&31.39	&63.97	&64.74	&\underline{16.79}\\
PEBEL~\cite{tian2022pixel}(ECCV'22)&49.14	&40.82	&4.98	&12.68	&44.17	&7.58	&\underline{92.38}	&\underline{1.73}	&45.1	&44.58	&47.15	&31.47\\
\hline
\textbf{Mask2Anomaly (Ours)}& \textbf{88.7}	&\underline{14.60}	&\textbf{93.3}	&\textbf{0.20}	&\underline{46.04}	&\underline{4.36}	&\textbf{95.20}	&\textbf{0.82}	&\underline{79.70}	&\underline{13.45}	&\textbf{80.59}	&\textbf{6.68}\\
\end{tabular}}
\caption{\textbf{Pixel level evaluation:} On average, Mask2Anomaly shows significant improvement  among the compared methods. Higher values for AuPRC are better, whereas for FPR$_{95}$ lower values are better. The best and second best results are \textbf{bold} and \underline{underlined}, respectively. `-' indicates the unavailability of benchmark results. } \vspace{-1em}
\label{tab:main} 
\end{table*}

\begin{table}[!ht]
\centering
\renewcommand{\arraystretch}{1.15}
\resizebox{1\linewidth}{!}{\begin{tabular}{c|ccc|ccc}
\multicolumn{1}{c|}{}&\multicolumn{3}{c|}{SMIYC RA-21}&\multicolumn{3}{c|}{SMIYC RO-21}\\
\cline{1-4}\cline{5-7}
\multicolumn{1}{c|}{Methods}&\multicolumn{1}{c|}{sIoU $\uparrow$}&\multicolumn{1}{c|}{PPV $\uparrow$} &\multicolumn{1}{c|}{$F1^{*}$$\uparrow$}&{sIoU $\uparrow$}&\multicolumn{1}{c|}{PPV $\uparrow$} &\multicolumn{1}{c|}{$F1^{*}$$\uparrow$}\\
\Xhline{3\arrayrulewidth}
Max Softmax~\cite{hendrycks2016baseline}(ICLR'17)  &			15.48 &	 	15.29 	 &	5.37 	 &	19.72  &		15.93  &		6.25 \\
Ensemble~\cite{lakshminarayanan2017simple}(NurIPS'17)&	16.44 	 &	20.77  &		3.39  &		8.63 	 &	4.71  &		1.28 \\
Mahalanobis~\cite{lee2018simple}(NeurIPS'18)   &		14.82  &		10.22 &	 	2.68  &		13.52 	 &	21.79 	 &	4.70  \\
Image Resynthesis~\cite{lis2019detecting}(ICCV'19) &	39.68	 &	10.95	 &	12.51	 &	16.61	 &	20.48 &		8.38	  \\
MC Dropout~\cite{mukhoti2018evaluating}(CVPR'20)   &	20.49 	 &	17.26 	 &	4.26 	 &	5.49 	 &	5.77 	 &	1.05 	 \\

Learning Embedding~\cite{blum2021fishyscapes}(IJCV'21)    &		33.86	 &	20.54 	 &	7.90  &			35.64  &		2.87  &		2.31  \\

SML~\cite{jung2021standardized}(ICCV'21)    &	26.00 &	 	24.70  &		12.20 	 &	5.10  &		13.30  &		3.00 \\

SynBoost~\cite{di2021pixel}(CVPR'21)  &		34.68	 &	17.81	 &	9.99	 &	44.28	 &	41.75	 &	37.57 \\
Maximized Entropy~\cite{chan2021entropy}(ICCV'21) &		49.21&	\underline{39.51}	&28.72	&\underline{47.87}	&\underline{62.64}	&48.51\\

JSRNet~\cite{vojir2021road}(ICCV'21) &		20.20  &		29.27 	 &	13.66 	 &	18.55 	 &	24.46  &		11.02 	\\
Void Classifier~\cite{blum2021fishyscapes}(IJCV'21)&	21.14 	 &	22.13  &		6.49  &		6.34  &		20.27  &		5.41  \\
Dense Hybrid~\cite{grcic2022densehybrid}(ECCV'22)&	\underline{54.17} &		24.13 &		\underline{31.08} &		45.74	 &	50.10	 &	\underline{50.72}	\\
PEBEL~\cite{tian2022pixel}(ECCV'22) &	38.88	 &	27.20	 &	14.48	 &	29.91	 &	 7.55 &		5.54	\\
\hline
Mask2Former~\cite{cheng2022masked}    &		25.20 	 &	18.20  &		15.30  &		5.00 	 &	21.90  &		4.80   \\
\textbf{Mask2Anomaly (Ours)}   &		\textbf{60.40}  &		\textbf{45.70}  &		\textbf{48.60}  &	 	\textbf{61.40}  &		\textbf{70.30}  &		\textbf{69.80} 	  \\
\end{tabular}}
\caption{\textbf{Component level evaluation:} Mask2Anomaly achieves large improvement on component level evaluation metrics among the baselined methods. Higher values of sIoU, PPV, and $F1^{*}$  are better. The best and second best results are \textbf{bold} and \underline{underlined}, respectively.
} 
\vspace{-1em}
\label{tab:comp-eval} 
\end{table}

\myparagraph{Evaluation Metrics:} We evaluate all the anomaly segmentation methods at pixel and component levels. For pixel-wise evaluation, we use Area under the Precision-Recall Curve (AuPRC) and False Positive Rate at a true positive rate of 95\% (FPR$_{95}$). Since pixel-level evaluation metrics can neglect small anomalies and be biased towards anomalies with large sizes, we also include component-level evaluations using the averaged component-wise F1 $(F1^{*})$, the positive predictive value (PPV), and the component-wise intersection over union (sIoU). Further, details of all the metrics can be found in the supplementary material.

\myparagraph{Implementation Details:}  Our implementation is  derived from~\cite{cheng2021per,cheng2022masked}. We use a ResNet-50~\cite{he2016deep} encoder, and its weights are initialized from a model that is pre-trained with barlow-twins~\cite{zbontar2021barlow} self-supervision on ImageNet~\cite{deng2009imagenet}. We freeze the encoder weights during training, saving memory and training time. We use a multi-scale deformable attention Transformer (MSDeformAttn)~\cite{zhu2020deformable} as the pixel decoder. The MSDeformAttn gives features maps at $1/8, 1/16,$ and $1/32$ resolution, providing image features to the transformer decoder layers. Our transformer decoder is adopted from~\cite{cheng2022masked} and consists of 9 layers with 100 queries. We train~\our{} using a combination of binary cross-entropy loss and the dice loss~\cite{milletari2016v} for class masks and cross-entropy loss for class scores. The network is trained with an initial learning rate of 1e-4 and batch size of 16 for 90 thousand iterations on AdamW~\cite{loshchilov2017decoupled} with a weight decay of 0.05. We use an image crop of $380\times760$ with large-scale jittering~\cite{du2021simple} along with a random scale ranging from 0.1 to 2.0.

Next, we train the~\our{} in a contrastive setting. We generate the outlier image using AnomalyMix~\cite{tian2022pixel} where we cut an object from MS-COCO~\cite{lin2014microsoft} dataset image and paste them on the Cityscapes image. The corresponding binary mask for an outlier image is created by assigning $1$ to the MS-COCO image area and $0$ to the Cityscapes image area. We randomly sample 300 images from the MS-COCO dataset during training to generate outliers. We train the network for 4000 iterations with $m$ as 0.75, a learning rate of 1e-5, and batch size 8, keeping all the other hyper-parameters the same as above. The probability of choosing an outlier in a training batch is kept at 0.2. 
\begin{figure}[t]
    \begin{center}
    \rotatebox{90}{\tiny{\hspace{2em}Ground Truth\hspace{4em}Mask2Anomaly\hspace{2.25em}Maximized Entropy~\cite{chan2021entropy}\hspace{1.25em}Dense Hybrid~\cite{grcic2022densehybrid}\hspace{3em}Input Image}}
     \includegraphics[width=0.4\textwidth, height=0.5\textwidth]{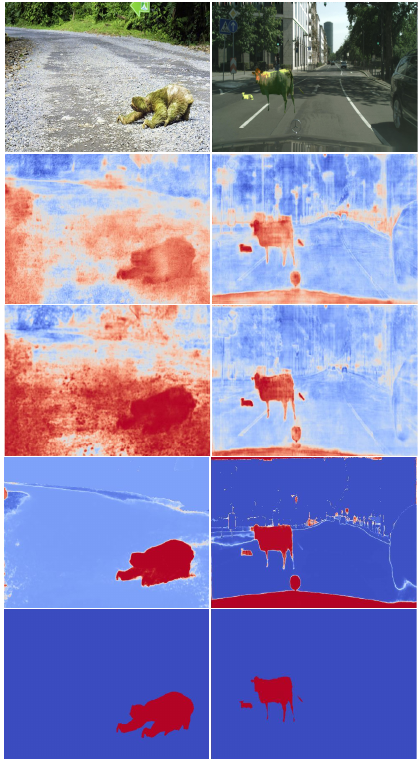}
    \end{center}
    \vspace{-1.25em}
\caption{\textbf{Qualitative Results}: We observe that per-pixel classification architectures: Dense Hybrid~\cite{grcic2022densehybrid} and Maximized Entropy~\cite{chan2021entropy} suffer from large false positives, whereas~\our{}, which is a mask-transformer, shows accurate pixel-wise anomaly segmentation results. }
    \label{fig:main} \vspace{-1em}
\end{figure}
\begin{figure*}[t]
\begin{center}
\includegraphics[width=0.95\linewidth, height=0.21\linewidth]{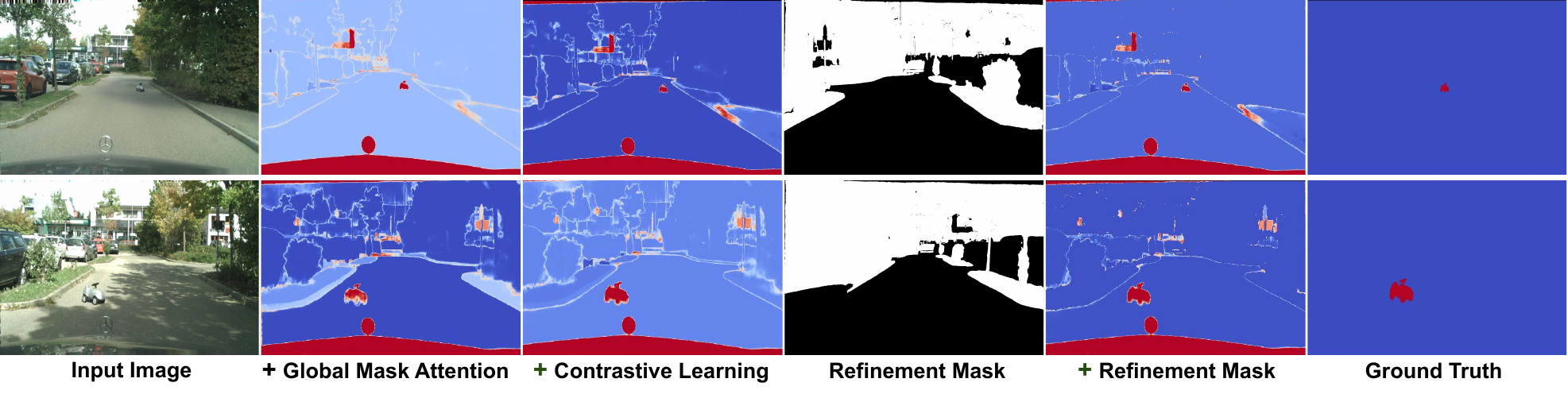}
\end{center}
\vspace{-1.5em}
  \caption{\textbf{Mask2Anomaly Qualitative Ablation}: 
demonstrates the performance gain by progressively adding (left to right ) proposed components. Masked-out regions by refinement mask are shown in white. Anomalies are represented in {red}.}
\label{fig:ablation} \vspace{-1em}
\end{figure*}

\subsection{Main Results}
\Cref{tab:main} shows the pixel-level anomaly segmentation results achieved by {\our} and recent SOTA methods on Fishyscapes, SMIYC, and Road Anomaly datasets. We can observe that~\our{} significantly improves the average AuPRC by 20\% and the FPR$_{95}$ by 60\% compared to the second-best method. 
We observe that anomaly segmentation methods based on per-pixel architecture, such as JSRNet, perform exceptionally well on the Road Anomaly dataset. However, JSRNet does not generalize well on other datasets. On the other hand,~\our{} yields excellent results on all the datasets.
Moreover, the property of our mask architecture to encourage objectness, rather than individual pixel anomalies, not only reduces the false positive but also improves the localization of whole anomalies. Indeed, \cref{tab:comp-eval} demonstrates that \our{}  outperforms all the baselined methods on component-level evaluation metrics. To conclude, {\our} yields state-of-the-art anomaly segmentation performance both in pixel and component metrics.

\myparagraph{Qualitative results:} To get a better understanding of the visual results, in \cref{fig:main} we visually compare the anomaly scores predicted by {\our} and its closest competitors: Dense Hybrid~\cite{grcic2022densehybrid} and Maximized Entropy~\cite{chan2021entropy}. 
The results from both: Dense Hybrid and Maximized Entropy exhibit a strong presence of false positives across the scene, particularly on the boundaries of objects (``things'') and regions (``stuff''). On the other hand, \our{} demonstrates the precise segmentation of anomalies while at the same time having minimal false positives. Additional qualitative results are in the supplementary material.

\myparagraph{Segmentation results}: Another critical characteristic of any anomaly segmentation method is that it should not disrupt the in-distribution classification performance, or else it would make the semantic segmentation model unusable. We find that adding only GMA to the base model leads to in-distribution accuracy of 80.45 on the validation set of Cityscapes. The final {\our} model maintains an in-distribution accuracy of 78.88 mIoU, which is still 1.46 points higher than the vanilla Mask2Former. Moreover, it is important to note that both~\our{} and Mask2Former are trained for 90k iterations, indicating that, although ~\our{} additionally attends to the background mask region, it shows convergence similar to Mask2Former. Extended quantitative and qualitative segmentation results with both~\our{} and Mask2Former are presented in the supplementary material.

\subsection{Ablations}
\label{sec:ablation}
All the results reported in this section are from the Fishyscapes lost and found validation dataset.

\myparagraph{Mask2Anomaly:} \Cref{tab:ab}(a) presents the results of a component-wise ablation of the technical novelties included in {\our}. We use Mask2Former as the baseline. As shown in the table, removing any individual component from {\our} drastically reduces the results, thus proving that their individual benefits are complimentary. In particular, we observe that the global masked attention has a big impact on the AuPRC and the contrastive learning is very important for the FPR$_{95}$. The mask refinement brings further improvements to both. \Cref{fig:ablation} visually demonstrates the positive effect of all the components.
\begin{table*}[t]
\renewcommand{\arraystretch}{1}
    \begin{subtable}{1\linewidth}
      \centering
        \begin{tabular}{ cc }   
        \resizebox{0.3\linewidth}{!}{\label{table-a}\begin{tabular}{ccc|cc}
        GMA & CL & RM & AuPRC$\uparrow$ & FPR$_{95}$$\downarrow$\\
        \Xhline{3\arrayrulewidth}
         &  &   &\textit{10.60} &\textit{89.35}  \\
        \hline
        \cmark  &  &  \cmark &35.05 &87.11  \\
         & \cmark & \cmark  &57.23 &31.93 \\ 
        \cmark  & \cmark &   &68.95 &24.07 \\ 
        \cmark  & \cmark &\cmark  &\textbf{69.41} &\textbf{9.46} \\ 
        \end{tabular}} &  
        \resizebox{0.25\linewidth}{!}{\begin{tabular}{c|cc}
        margin($m$) &  AuPRC$\uparrow$ & FPR$_{95}$$\downarrow$\\
        \Xhline{3\arrayrulewidth}
        $1$   &65.37 &11.61 \\ 
        $0.95$ &65.40 &12.20\\ 
        $0.90$ &66.05 &13.49\\
        $0.80$ &66.20 &14.89\\
        $0.75$ &\textbf{69.41} &\textbf{9.46}\\ 
        $0.50$ &62.07 &13.26\\
        \end{tabular}}\\
        (a) & (b) 
        \\ 
        \end{tabular}
    \end{subtable}
    
    \begin{subtable}{1\linewidth}
      \centering
        \begin{tabular}{ ccc }   
        \resizebox{0.3\linewidth}{!}{\begin{tabular}{cccc}
        & mIoU$\uparrow$ & AuPRC$\uparrow$ & FPR$_{95}$$\downarrow$\\
        \Xhline{3\arrayrulewidth}
        CA~\cite{cheng2021per}& 76.43  &20.30 &89.35  \\
        MA~\cite{cheng2022masked}& 77.42&10.60 &89.39  \\ 
        \cellcolor{blue!15}{GMA} &\textbf{80.45} &\textbf{32.35} &\textbf{25.95} \\ 
        \end{tabular}} &  
        \resizebox{0.32\linewidth}{!}{\begin{tabular}{c|cc}
        & AuPRC$\uparrow$ & FPR$_{95}$$\downarrow$\\
        \Xhline{3\arrayrulewidth}
        $w/o$ Refinement Mask  &68.95 &24.07 \\
        \hline
        $L_{\{things \hspace{0.1em} \setminus \hspace{0.1em} road\}}$  &67.04 &39.11 \\ 
        
        \cellcolor{blue!15}\textbf{$L_{\{stuff \hspace{0.1em} \setminus \hspace{0.1em} road\}}$}  &\textbf{69.41} &\textbf{9.46} \\ 
        \end{tabular}} &
        
        \resizebox{0.3\linewidth}{!}{\begin{tabular}{c|cc}
        Batch Outlier Probability & AuPRC$\uparrow$& FPR$_{95}$$\downarrow$\\
        \Xhline{3\arrayrulewidth}
        0.1  &63.01 &14.66 \\
        0.2  &\textbf{69.41} &\textbf{9.46}  \\ 
        0.5 &69.20 &11.03 \\ 
        1  &68.77 &10.53 \\ 
        \end{tabular}} \\
        (c) & (d) & (e)\\ 
        \end{tabular}
    \end{subtable}
    \caption{\textbf{\our{} Ablation tables:} \textbf{(a)} Component-wise ablation of Mask2Anomaly. Results in \textit{italics} show Mask2Former results. GMA: Global Mask Attention, CL: Contrastive Learning, and RM: Refinement Mask. \textbf{(b)} Shows the behavior of $L_{CL}$ by choosing different margin($m$) values. We empirically find the best results when $m$ is 0.75.  \textbf{(c)} Global masked attention (GMA) performs the best among various attention mechanisms: Cross-Attention (CA) and Masked-Attention (MA). \textbf{(d)} We show the performance gain by using a refinement mask that masks the $ \{stuff \setminus road\} $ regions as anomalies are categorized as $things$ class.  \textbf{(e)} Batch outlier probability is the likelihood of selecting an outlier image for a batch during contrastive training. The best result is achieved at 0.2 probability.    (\textit{All the results reported on FS Lost \& Found validation set}).}
\label{tab:ab}
\end{table*}

\begin{figure}[t]
\begin{center}
\rotatebox{90}{\tiny{\hspace{3.5em}CA~\cite{cheng2021per}\hspace{3.5em}MA~\cite{cheng2022masked}\hspace{2em}GMA (Ours)}}
\includegraphics[width=0.95\linewidth, height=0.45\linewidth]{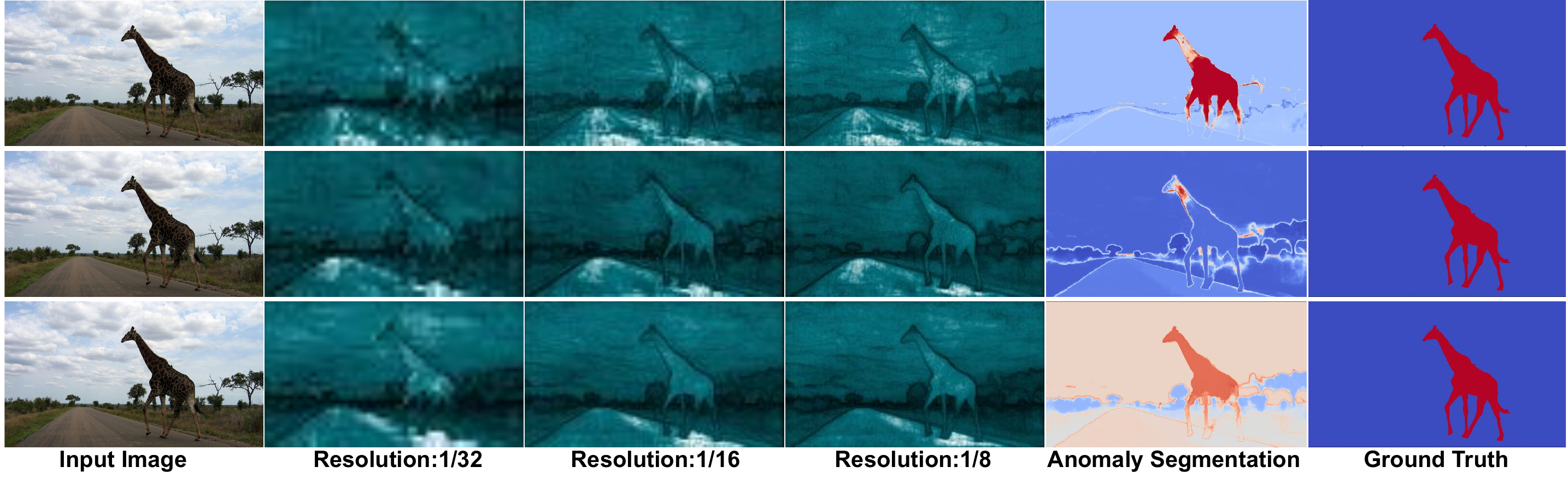}
\end{center} \vspace{-1em}
 \caption{\textbf{Visualization of negative attention maps and results:} Global mask attention gives high attention scores to anomalous regions across all resolutions showing the best anomaly segmentation results among the compared attention mechanisms. Cross-attention performs better than mask-attention but has high false positives and low confidence prediction for the anomalous region. Darker regions represent low attention values. Details to calculate negative attention are given in Section:\ref{sec:ablation}.}
\label{fig:attention}
\end{figure}
\myparagraph{Global Mask Attention:} To better understand the effect of the global masked attention (GMA), in \cref{tab:ab}(c), we compare it to the masked-attention (MA)~\cite{cheng2022masked} and cross-attention (CA)~\cite{vaswani2017attention}. We can observe that although the MA increases the mIoU w.r.t. the CA, it degrades all the metrics for anomaly segmentation, thus confirming our preliminary experiment shown in \cref{fig:mask2former_attention}. On the other hand, the GMA provides improvements across all the metrics.
This is confirmed visually in \cref{fig:attention}, where we show the negative attention maps for the three methods at different resolutions.
The negative attention is calculated by averaging all the queries (since there is no reference known object) and then subtracting one. Note that the GMA has a high response on the anomaly (the giraffe) across all resolutions.

\myparagraph{Refinement Mask:} \Cref{tab:ab}(d) shows the performance gains due to the refinement mask. We observe that filtering out the $\{ \text{``stuff''} \setminus \text{``road''} \} $ regions of the prediction map improves the FPR$_{95}$ by $14.61$ along with marginal improvement in AuPRC. On the other hand, removing the $\{ \text{``things''} \setminus \text{``road''}\}$ regions degrades the results, confirming our hypothesis that anomalies are likely to belong to the ``things'' category. \Cref{fig:ablation} qualitatively shows the improvement achieved with the refinement mask. Also, refinement mask adds a small overhead of 1.12 GFlops compared to Mask2Anomaly 258 GFlops inference cost.

\myparagraph{Mask Contrastive Learning:} We tested the effect of the margin in the contrastive loss $L_{CL}$, and we report these results in \cref{tab:ab}(b). We find that the best results are achieved by setting $m$ to 0.75, but the performance is competitive for any value of $m$ in the table. Similarly, we tested the effect of the batch outlier probability, which is the likelihood of selecting an outlier image in a batch. The results shown in \cref{tab:ab}(e) indicate that the best performance is achieved at $0.2$, but the results remain stable for higher values of the batch outlier probability.
\begin{table}[t]
\centering
\renewcommand{\arraystretch}{1.15}
\resizebox{1\linewidth}{!}{\begin{tabular}{c|c|ccc|cc}
\multirow{2}{*}{Method} &\multirow{2}{*}{Backbone} &\multirow{2}{*}{AuPRC$\uparrow$} &\multirow{2}{*}{FPR$_{95}\downarrow$}  & \multirow{2}{*}{FLOPs$\downarrow$} & Training $\downarrow$\\\
& & & & & Parameters \\
\Xhline{3\arrayrulewidth}
Mask2Former~\cite{cheng2022masked}& ResNet-50 &10.60 &89.35 &\textbf{226G} &44M\\

& ResNet-101 &9.11 &45.83 &293G &63M\\

& Swin-T &24.54 &37.98 &232G &42M\\
& Swin-S &30.96 &36.78 &313G &69M\\
\hline
Mask2Anomaly$^\ddag$ & ResNet-50 &\textbf{32.35} &\textbf{25.95} &258G &\textbf{23M}\\
\end{tabular}}
\caption{\textbf{Architectural Efficiency of Mask2Anomaly:} Mask2Anomaly outperforms the best Mask2Former architecture having Swin-S backbone with only 30\% trainable parameters. Mask2Anomaly$^\ddag$ only uses global mask attention.}
\label{tab:largebackbone} \vspace{-1em}
\end{table}

\myparagraph{Effect of bigger backbones:} We demonstrate the efficacy of {\our} by comparing it to the vanilla Mask2Former but using larger backbones. The results in~\cref{tab:largebackbone} show that despite the disadvantage, Mask2Anomaly with a ResNet-50 still performs better than Mask2Former using large transformer-based backbones. It is also important to note that the number of training parameters for Mask2Anomaly can be reduced to $23M$ by using a frozen self-supervised pre-trained encoder, which is significantly less than all the Mask2Former variations. 

\section{Conclusion}
In this work, we present Mask2Anomaly, a novel anomaly segmentation architecture established on masked architecture. Mask2Anomaly contains global mask attention specifically designed to improve the attention mechanism for anomaly segmentation tasks. Next, we develop a mask contrastive learning framework that utilizes outlier masks to maximize the separation between anomalies and known classes. Finally, 
we introduced mask refinement that reduces false positives and improves the overall performance. We show the efficacy of Mask2Anomaly and its components through extensive qualitative and quantitative results. We hope Mask2Anomaly will open doors for new anomaly segmentation methods based on mask architecture.

{\small
\bibliographystyle{ieee_fullname}
\bibliography{egbib}
}

\clearpage
\noindent{\large\textbf{Supplementary Material}}

\paragraph{Summary:} This supplementary material contains additional method explanations, experiments, and results of~\our{} that include:
\begin{itemize}[noitemsep,topsep=0pt]
    \item explanation of anomaly segmentation evaluation metrics;
    \item\our{} results on validation sets;
    \item outlier loss comparison and analysis;
    \item training loss functions of~\our{} ;
    \item an analysis of various inference techniques applied to a {\our};
    \item performance stability of~\our{};
    \item additional results and supplementary video.
\end{itemize}

\myparagraph{A. Evaluation Metrics}\\
\textbf{Pixel-Level:}  For pixel-wise evaluation, consider $Y \in \{ Y_{a}, Y_{na}\}$ is the pixel level annotated ground truth labels for image $\chi$ containing anomaly. $Y_{a}$ and $Y_{na}$ represents the anomalous and non-anomalous labels in the ground-truth. Assume, $\hat{Y}(\gamma)$ is the model prediction obtained at thresholding $f(x)$ at $\gamma$. Then, we can write precision and recall equations as:
\begin{equation}
\text{precision}(\gamma) = \frac{|Y_{a} \cap \hat{Y}_{a}(\gamma)|} {|\hat{Y}_{a}(\gamma)|}
\end{equation}

\begin{equation}
\text{recall}(\gamma) = \frac{|Y_{a} \cap \hat{Y}_{a}(\gamma)|}{|Y_{a}|}
\end{equation}

\noindent and, AuPRC can be approximated as:

\begin{equation}
\text{AuPRC} = \int_{\gamma}^{}\text{precision}(\gamma)\text{recall}(\gamma)
\end{equation}
The AuPRC works well for unbalanced datasets making it particularly suitable for anomaly segmentation since all the datasets are significantly skewed. Next, we consider the False Positive Rate at a true positive rate of 95\% (FPR$_{95}$), an important criterion for safety-critical applications that is calculated as:
\begin{equation}
\text{FPR}_{95} = \frac{|\hat{Y}_{a}(\gamma^{*}) \cap Y_{na}|}{|Y_{na}|}
\end{equation}

where $\gamma^{*}$ is a threshold when the true positive rate is 95\%.

\noindent\textbf{Component-Level:} SMIYC~\cite{chan2021segmentmeifyoucan} introduced a few component-level evaluation metrics that solely focus on detecting anomalous objects regardless of their size. These metrics are important to be considered because pixel-level metrics may not penalize a model for missing a small anomaly, even though such a small anomaly may be important to be detected. In order to have a component-level assessment of the detected anomalies, the quantities to be considered are the component-wise true-positives ($TP$), false-negatives ($FN$), and false-positives ($FP$). These component-wise quantities can be measured by considering the anomalies as the positive class. From these quantities, we can use three metrics to evaluate the component-wise segmentation of anomalies: sIoU, PPV, and F1$^{*}$. Here we provide the details of how these metrics are computed, using the notation $\mathcal{K}$ to denote the set of ground truth components, and $\hat{\mathcal{K}}$ to denote the set of predicted components.

The \textit{sIoU} metric used in SMIYC~\cite{chan2021segmentmeifyoucan} is a modified version of the component-wise intersection over union proposed in~\cite{rottmann2020prediction}, which considers the ground-truth components in the computation of the $TP$ and $FN$. Namely, it is computed as 
\begin{equation}
    \text{sIoU}(k) = \frac{|k\cap \hat{K}(k)|}{|k\cap \hat{K}(k) \backslash \mathcal{A}(k)|},
    \qquad \hat{K}(k) = \bigcup_{\hat{k} \in \hat{\mathcal{K}}, \, \hat{k} \cap k \neq \emptyset} \hat{k}
\end{equation}
where $\mathcal{A}(k)$ is an adjustment term that excludes from the union those pixels that correctly intersect with another ground-truth component different from $k$. We refer the reader to~\cite{chan2021segmentmeifyoucan} for more details on this term.
Given a threshold $\tau \in [0, 1]$, a target $k \in \mathcal{K}$ is considered a $TP$ if $sIoU(k) > \tau$, and a $FN$ otherwise.

The positive predictive value (\textit{PPV}) is a metric that measures the $FP$ for a predicted component $\hat{k} \in \hat{\mathcal{K}}$, and it is computed as 
\begin{equation}
    \text{PPV}(\hat{k}) = \frac{|\hat{k}\cap \hat{K}(k)|}{|\hat{k}|}
\end{equation}
A predicted component $\hat{k} \in \hat{\mathcal{K}}$ is considered a $FP$ if $PPV(\hat{k}) \leq \tau$.
\begin{table}[t]
\centering
\renewcommand{\arraystretch}{1}
\resizebox{1\linewidth}{!}{\begin{tabular}{c|cc|cc}
\multicolumn{1}{c|}{}&\multicolumn{2}{c|}{FS   L\&F}&\multicolumn{2}{c|}{FS static}\\
\cline{2-3}\cline{4-5}
\multicolumn{1}{c|}{Methods}&\multicolumn{1}{c|}{AuPRC$\uparrow$}&{FPR$_{95}$ $\downarrow$}&\multicolumn{1}{c|}{AuPRC$\uparrow$}&{FPR$_{95}$ $\downarrow$}\\
\Xhline{3\arrayrulewidth}
Max Softmax~\cite{hendrycks2016baseline}  &			4.59 &	 	40.59 	 &			19.09  &		23.99 \\
Max Logit~\cite{hendrycks2016baseline}  &			14.59 &	 	42.21 	 &			38.64  &		18.26 \\
Entropy~\cite{hendrycks2016baseline}  &			10.36 &	 	40.34 	 &			26.77  &		23.31 \\
Energy~\cite{liu2020energy}  &			25.79 &	 	32.26 	 &			31.66  &		37.32 \\
SynthCP~\cite{xia2020synthesize}  &		6.54 &	45.95	 &	23.22		 &	34.02 \\
SynBoost~\cite{di2021pixel}  &		40.99	 &	34.47	 &	48.44		 &	47.71 \\
SML~\cite{jung2021standardized}   &	36.55 &	 	14.53  &			48.67 &		16.75 \\
Deep Gambler~\cite{liu2019deep}   &	39.77 &	 	12.41  &			67.69  &		15.39\\
Dense Hybrid~\cite{grcic2022densehybrid}&		\underline{63.80} &	\textbf{6.10}	 &	60.20	 &	\underline{4.90}	\\
PEBEL~\cite{tian2022pixel} &	59.83	 &	\underline{6.49}	 &	 \underline{82.73} &		6.81	\\
\hline
\textbf{Mask2Anomaly (Ours)} &		\textbf{69.41}  &	 	9.46  &		\textbf{90.54}  &		\textbf{1.98} 	  \\

\end{tabular}}
\caption{\textbf{Fishyscapes Validation Results:} The best and second best results are \textbf{bold} and \underline{underlined}, respectively.
} 
\label{tab:fs-val} 
\end{table}
Finally, the \textit{$F1^{*}$} summarizes all the component-wise $TP$, $FN$, and $FP$ quantities by the following formula:
\begin{equation}
    \label{f1*}
    F1^{*}(\tau) = \frac{2TP(\tau)}{2TP(\tau) + FN(\tau) + FP(\tau)}
\end{equation}
 
\myparagraph{B. Results on Fishyscapes and SMIYC validation sets}\\
To provide a comprehensive evaluation, we have benchmarked~\our{} results on the Fishyscapes and SMIYC validation sets as presented in~\cref{tab:fs-val} and~\cref{tab:smiyc-val}, respectively. We can observe that~\our{} outperforms all the prior methods by a large margin on both benchmarks. Interestingly, maximized entropy and dense hybrid show the best AuPRC for SMIYC-RO21 and FPR$_{95}$ for FS L\&F, respectively. However, overall~\our{} gives the best performance on all the benchmarks. This suggests that mask-based architecture offers better generalizability in comparison to per-pixel architecture due to its intrinsic property of encouraging objectness. \\

\begin{figure}[t]
    \begin{center}
        \includegraphics[width=1\linewidth]{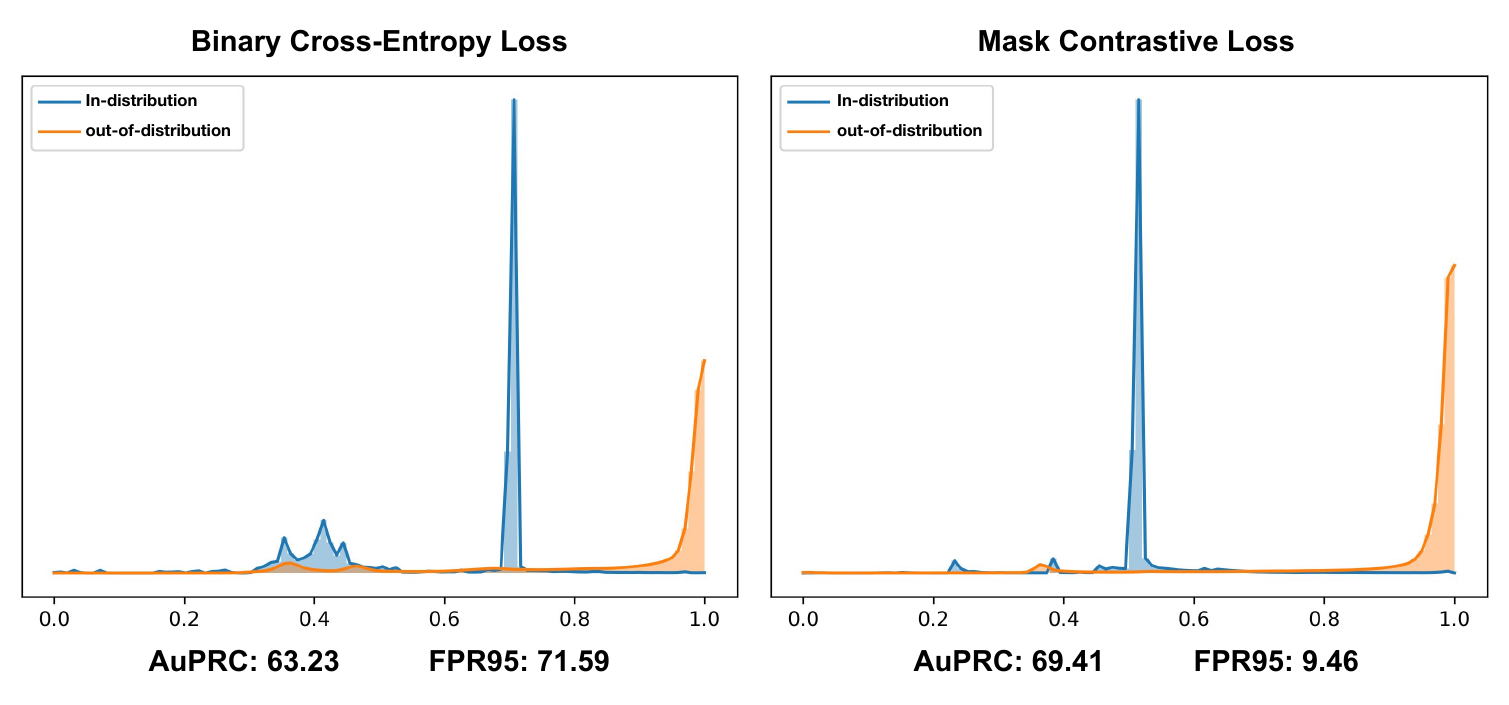}
    \end{center}
    \vspace{-1.5em}
\caption{\textbf{Outlier Loss Comparision:} To train~\our{} on the outlier set, we find that mask contrastive loss, which is a margin-based loss shows better performance compared to the binary cross-entropy loss. Both experiments are done on the FS L\&F validation set.}
    \label{fig:BCEvsCL}
    \vspace{-1em}
\end{figure}

\myparagraph{C. Outlier Loss Comparision}\\
We now empirically demonstrate why mask contrastive loss, a margin-based loss, performs better at anomaly segmentation than binary cross-entropy loss. We train~\our{} with $M_{OOD}$ using binary-cross entropy. The new loss based on the binary cross entropy can be written as:
\begin{equation}
L_{BCE}=M_{OOD}\log(l_{N})+(1-M_{OOD})\log(1-l_{N})
\end{equation}
\begin{equation}
    \begin{aligned}
         \text{where, }l_{N} = - \max_{k=1}^{K} \left(\text{softmax}(C)^T \cdot \text{sigmoid}(M)\right)
    \end{aligned}
\end{equation}
$l_{N}$ is the negative likelihood of in-distribution classes calculated using the class scores $C$ and class masks $M$.~\Cref{fig:BCEvsCL} illustrates the anomaly segmentation performance comparison on FS L\&F validation dataset between the~\our{} when trained with the binary cross entropy loss and mask contrastive loss, respectively. We can observe that the mask contrastive loss achieves a wider margin between out-of-distribution(anomaly) and in-distribution prediction while maintaining significantly lower false positives.\\
\begin{table}[t]
\centering
\renewcommand{\arraystretch}{1}
\resizebox{1\linewidth}{!}{\begin{tabular}{c|cc|cc}
\multicolumn{1}{c|}{}&\multicolumn{2}{c|}{SMIYC-RA21}&\multicolumn{2}{c|}{SMIYC-RO21}\\
\cline{2-3}\cline{4-5}
\multicolumn{1}{c|}{Methods}&\multicolumn{1}{c|}{AuPRC$\uparrow$}&{FPR$_{95}$ $\downarrow$}&\multicolumn{1}{c|}{AuPRC$\uparrow$}&{FPR$_{95}$ $\downarrow$}\\
\Xhline{3\arrayrulewidth}
Max Softmax~\cite{hendrycks2016baseline}  &			40.4 &	 	60.2 	 &			43.4  &		3.8 \\
ODIN~\cite{liang2017enhancing}  &			46.3 &	 	61.5 	 &			46.6  &		4.0 \\
Mahalanobis~\cite{lee2018simple}   &		22.5  &		86.4  	 &	25.9 	 &	26.1  \\
MC Dropout~\cite{mukhoti2018evaluating}  &	29.2	 &	77.9	 &	7.9	 &	43.8 	 \\
Ensemble~\cite{lakshminarayanan2017simple} &	16.0 	 &	80.0  &		4.7  &		98.3 \\
Void Classifier~\cite{blum2021fishyscapes}  &		39.3  &		66.1  &		9.8  &		43.6  \\
Learning Embedding~\cite{blum2021fishyscapes}   &		51.9	 &			60.0  &		1.5  &		56.7  \\
Image Resynthesis~\cite{lis2019detecting} &	76.4	 &	20.5		 &	70.3 &		1.3	  \\
SynBoost~\cite{di2021pixel}  &		68.8	 &	30.9	 &	81.4	 &	2.8 \\
Maximized Entropy~\cite{chan2021entropy}	&\underline{80.7}	&\underline{17.4}	&\textbf{94.4}	&\underline{0.4}\\
\hline
\textbf{Mask2Anomaly (Ours)} &		\textbf{94.5}  &	 	\textbf{3.3}  &		\underline{88.6}  &		\textbf{0.3} 	  \\

\end{tabular}}
\caption{\textbf{SMIYC Validation Results:} The best and second best results are \textbf{bold} and \underline{underlined}, respectively.
} 
\label{tab:smiyc-val} 
\end{table}
\begin{table*}[!ht]
\centering
\renewcommand{\arraystretch}{1.1}
\resizebox{1\linewidth}{!}{\begin{tabular}{ccc|cc|cc|cc|cc|cc|cc}
\multicolumn{3}{c|}{}&\multicolumn{2}{c|}{SMIYC-RA21}&\multicolumn{2}{c|}{SMIYC-RO21}&\multicolumn{2}{c|}{FS L\&F}&\multicolumn{2}{c|}{FS Static}&\multicolumn{2}{c|}{Road Anomaly}& \multicolumn{2}{c}{Average}\\
\cline{4-15}
\multicolumn{1}{c|}{$C$}&\multicolumn{1}{c|}{$M$}&\multicolumn{1}{c|}{$f(C).f(M)$}&\multicolumn{1}{c|}{AuPRC $\uparrow$} &\multicolumn{1}{c|}{FPR$_{95}$ $\downarrow$}&\multicolumn{1}{c|}{AuPRC $\uparrow$} &\multicolumn{1}{c|}{FPR$_{95}$ $\downarrow$}&\multicolumn{1}{c|}{AuPRC $\uparrow$} &\multicolumn{1}{c|}{FPR$_{95}$ $\downarrow$}&\multicolumn{1}{c|}{AuPRC $\uparrow$} &\multicolumn{1}{c|}{FPR$_{95}$ $\downarrow$}&\multicolumn{1}{c|}{AuPRC $\uparrow$} &\multicolumn{1}{c|}{FPR$_{95}$ $\downarrow$}&\multicolumn{1}{c|}{AuPRC $\uparrow$} &\multicolumn{1}{c|}{FPR$_{95}$ $\downarrow$}\\

\Xhline{3\arrayrulewidth}

$I$	&$I$	&$I$ &9.47	&95.16	&4.44	&73.45	&2.53	&92.16	&1.18	&99.97	&65.59	&97.56	&16.64	&91.66	\\
Softmax	&Softmax	&$I$	&44.73	&38.27	&3.16	&95.72	&4.82	&47.98	&10.34	&52.04	&42.74	&55.73	&21.13	&57.94\\
Sigmoid	&Sigmoid	&$I$	&25.04 &93.14 &83.14 &1.24 &14.55 &43.83 &45.67 &96.87 &28.1 &91.63 &39.3 &65.34\\
Sigmoid 	&Softmax	&$I$	&29.29 &39.01 &7.48 &98.01 &0.42 &48.23 &6.37 &52.16 &25.61 &55.78 &13.83 &58.63\\
Softmax	&Sigmoid	&$I$	&95.48 &2.41 &92.89 &0.15 &69.41 &9.46 &90.54 &1.98 &79.7 &13.45 &\textbf{85.56} &\textbf{5.51}
\\
Softmax	&Sigmoid	&Softmax	&94.55 &3.31 &88.59 &0.36 &70.8 &32.66 &88.96 &2.22 &78.3 &15.54 &84.24 &10.81\\

\end{tabular}}
\caption{\textbf{Mask2Anomaly Inference}: we show various inference techniques on Mask2Anomaly for anomaly segmentation. $f(.)$ represents the function applied to class scores or masks. $I$ is the identity function. The best results are in bold.} 
\label{tab:sup:inference} 
\end{table*}
\begin{table*}[ht]
\centering
\setlength{\tabcolsep}{2pt} 
\renewcommand{\arraystretch}{1.25}

\resizebox{0.8\linewidth}{!}{\begin{tabular}{c|cc|cc|cc|cc|cc}
\multicolumn{1}{c|}{}&\multicolumn{2}{c|}{SMIYC-RA21}&\multicolumn{2}{c|}{SMIYC-RO21}&\multicolumn{2}{c|}{FS L\&F}&\multicolumn{2}{c|}{FS Static}& \multicolumn{2}{c}{Average $\sigma$}\\
\cline{2-3}\cline{4-5}\cline{6-7}\cline{8-9}\cline{10-11}
\multicolumn{1}{c|}{Methods}&\multicolumn{1}{c|}{AuPRC $\uparrow$} &\multicolumn{1}{c|}{FPR$_{95}$ $\downarrow$}&\multicolumn{1}{c|}{AuPRC $\uparrow$} &\multicolumn{1}{c|}{FPR$_{95}$ $\downarrow$}&\multicolumn{1}{c|}{AuPRC $\uparrow$} &\multicolumn{1}{c|}{FPR$_{95}$ $\downarrow$}&\multicolumn{1}{c|}{AuPRC $\uparrow$} &\multicolumn{1}{c|}{FPR$_{95}$ $\downarrow$}&\multicolumn{1}{c|}{AuPRC } &\multicolumn{1}{c}{FPR$_{95}$}\\
\Xhline{3\arrayrulewidth}
Mask2Anomaly-S1  &95.48 &2.41 &92.89 &0.15 &69.41 &9.46 &90.54 &1.98 &- &- \\
Mask2Anomaly-S2  &92.03 &3.22 &92.3 &0.27 &69.19 &13.47 &85.63 &5.06 &- &- \\
$\sigma$(Mask2Anomaly)  &$\pm$ 2.44 &$\pm$0.57 &$\pm$0.42 &$\pm$0.08 &$\pm$0.16 &$\pm$2.84 &$\pm$3.47 &$\pm$2.18 &$\pm$1.62	&$\pm$1.41 \\
\hline

Dense Hybrid-S1 &52.99 &38.87 &66.91 &1.91 &56.89 &8.92 &52.58 &6.03 &- &-\\

Dense Hybrid-S2 &60.59 &32.14 &79.64 &1.01 &47.97 &18.35 &54.22 &5.24 &- &-\\

$\sigma$(Dense Hybrid) &$\pm$5.37 &$\pm$4.76 &$\pm$9.00 &$\pm$0.64 &$\pm$6.31 &$\pm$6.67 &$\pm$1.16 &$\pm$0.56 &$\pm$5.46	&$\pm$3.15
\end{tabular}}
\caption{\textbf{Performance stability in Mask2Former:} we can observe that the average deviation in the performance of the dense hybrid is significantly higher than Mask2Anomaly. $\sigma$ denotes the standard deviation. } \vspace{-1em}
\label{sup-tab:ood-sets} 
\end{table*}
\begin{table*}[t]
\centering
\renewcommand{\arraystretch}{1}
\resizebox{\linewidth}{!}{\begin{tabular}{ccccccccccccccccccccc}
\multicolumn{1}{c|}{Methods}&\multicolumn{1}{c|}{road}&\multicolumn{1}{c|}{s. walk}&\multicolumn{1}{c|}{building}&\multicolumn{1}{c|}{wall} &\multicolumn{1}{c|}{fence}&\multicolumn{1}{c|}{pole} &\multicolumn{1}{c|}{t. light}&\multicolumn{1}{c|}{t. sign} &\multicolumn{1}{c|}{veg.}&\multicolumn{1}{c|}{terrain} &\multicolumn{1}{c|}{sky}&\multicolumn{1}{c|}{person} &\multicolumn{1}{c}{rider}&\multicolumn{1}{c}{car}&\multicolumn{1}{c}{truck}&\multicolumn{1}{c}{bus}&\multicolumn{1}{c}{train}&\multicolumn{1}{c}{mbike}&\multicolumn{1}{c}{bicycle}&\multicolumn{1}{c}{mIoU}\\
\Xhline{3\arrayrulewidth}
Mask2Former & 98.4 &87.0 &92.7 &46.1 &59.9 &69.5 &75.3
&82.2 &92.9 &63.8 &95.2 &84.9 &69.3 &95.6 &58.7 &77.0 &79.9 &62.7 &80.0 &77.4\\
Mask2Anomaly &98.5 &86.3 &91.5 &53.9 &60.2 &67.5 &74.3 &88.1 &93.1 &62.6 &96 &84.1 &62.7 &95.7 &79.6 &80.3 &77.1 &70.1 &77.1 &\textbf{78.8}\\
\end{tabular}}
\caption{Class-wise semantic segmentation results comparison between Mask2Former and Mask2Anomaly on Cityscapes validation set.} \vspace{-1em}
\label{sstab:main} 
\end{table*}

\myparagraph{D. Training Loss}\\
\our{} gives two sets of outputs: class scores ($C$) and class masks ($M$). To train $M$, we first pad the ground truth mask $M^{gt}$ with “no object” masks denoted by $\phi$. Since we assume $M\geq M^{gt}$, padding the ground truth masks allow us one-to-one matching. Now, we use bipartite matching to match the ground truth and the predicted masks, and the assignment cost is given by:
\begin{equation}
    L_{masks} = \lambda_{bce}L_{bce} + \lambda_{dice}L_{dice}
\end{equation}
where $L_{bce}$ and $L_{dice}$ are the binary cross entropy loss and the dice loss calculated between the matched masks. $\lambda_{bce}$ and $\lambda_{dice}$ are the loss weights that are both set to $5.0$. 
To train $C$, which indicates the semantic class of a mask, we used the cross-entropy loss $L_{ce}$. The total training loss is given by:
\begin{equation}
    L = L_{masks} + \lambda_{ce}L_{ce}
\end{equation}
with $\lambda_{ce}$ set to 2.0 for the prediction that matched with ground truth and 0.1 for $\phi$, \ie for no object. After training the~\our{} for 90K iterations, we fine-tune the network with the mask contrastive loss $L_{CL}$. The new training loss is written as:
\begin{equation}
    L_{M2A} = L + L_{CL}
\end{equation}
We perform all the training and inference on a single Nvidia Titan RTX with 24GB memory.\\

\myparagraph{E. Mask2Anomaly Inference}\\
The per-pixel classification networks have a straightforward inference as the network outputs a pixel-wise anomaly map. However, in the case of a mask architecture, we get a set of class scores $C$ and a set of binary mask $M$. So, we test various inference techniques on Mask2Anomaly for anomaly segmentation, as shown in~\Cref{tab:sup:inference}. We find that the marginalization over class scores obtained after the softmax and taking the sigmoid of the mask yields the best results. Also, we observe that applying a softmax after the marginalization to perform max-softmax~\cite{hendrycks2016baseline} does not give good results.
\begin{figure*}[t]
    \begin{center}
        \includegraphics[width=1\linewidth,height=0.85\textwidth]{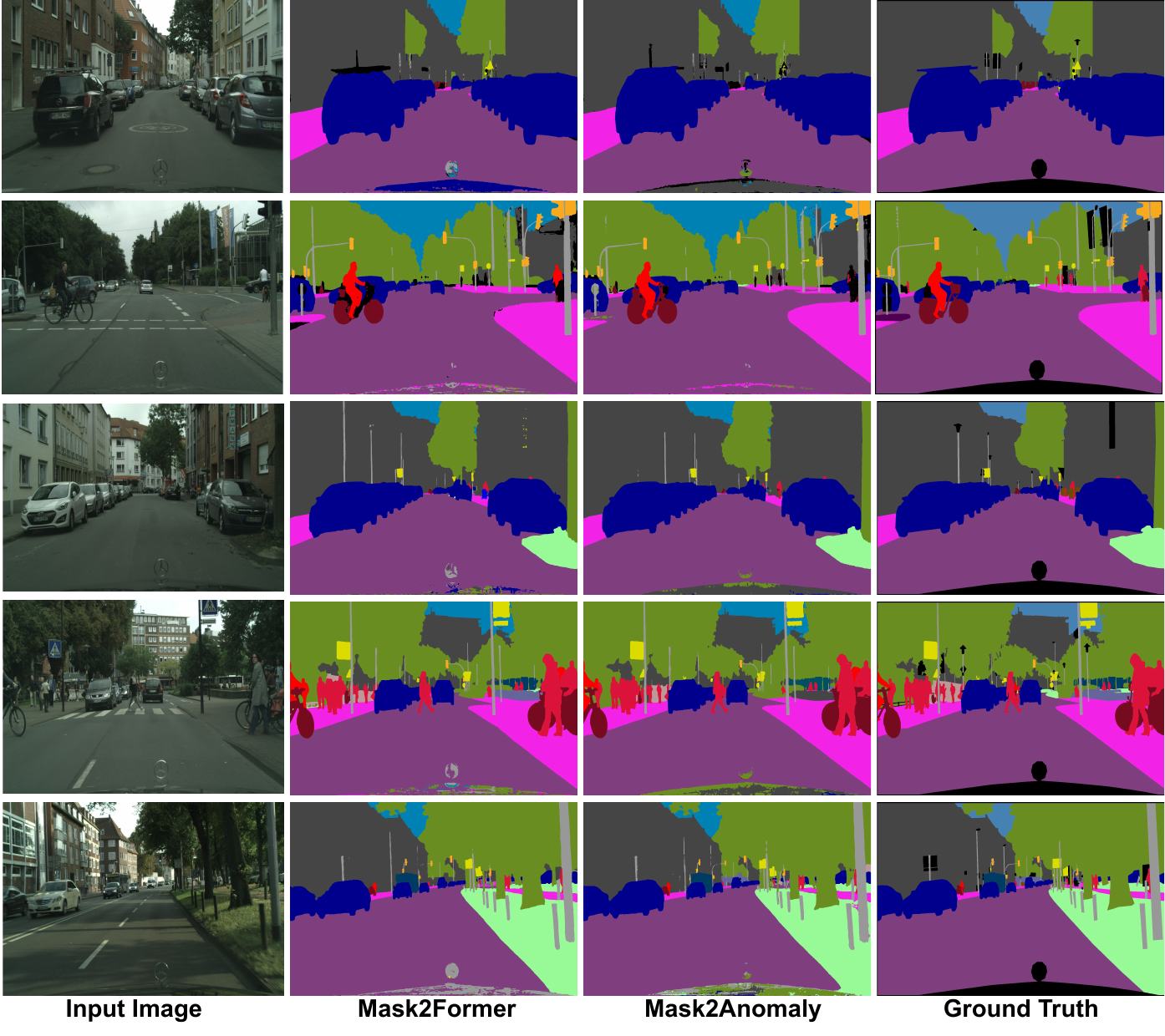}
    \end{center}
    \vspace{-1.5em}
\caption{\textbf{Semantic Segmentation Results:} We can visually infer that Mask2Anomaly shows similar segmentation results when compared with Mask2Former~\cite{cheng2022masked}.}
    \label{fig:ss}
    \vspace{-1em}
\end{figure*}
\\

\myparagraph{F. Performance stability on different outlier sets}\\
Employing an outlier set to train an anomaly segmentation model presents a challenge because the model's performance can vary significantly across different sets of outliers. Here, we show that~\our{} performs similarly when trained on different outlier sets.

We randomly chose two subsets of 300 MS-COCO images (S1, S2) as our outlier dataset for training~\our{} and DenseHybrid.~\Cref{sup-tab:ood-sets} shows the performance of~\our{} and Dense Hybrid trained on S1 and S2 outlier sets, along with the standard deviation($\sigma$) in the performance. We can observe that the variation in performance for the dense hybrid is significantly higher than~\our{}. Specifically, in dense hybrid, the average deviation in AuPRC is greater than 300\%, and the average variation in FPR$_{95}$ is more than 200\% compared to~\our{}. 

\begin{figure*}[t]
    \begin{center}
        \includegraphics[width=1\textwidth]{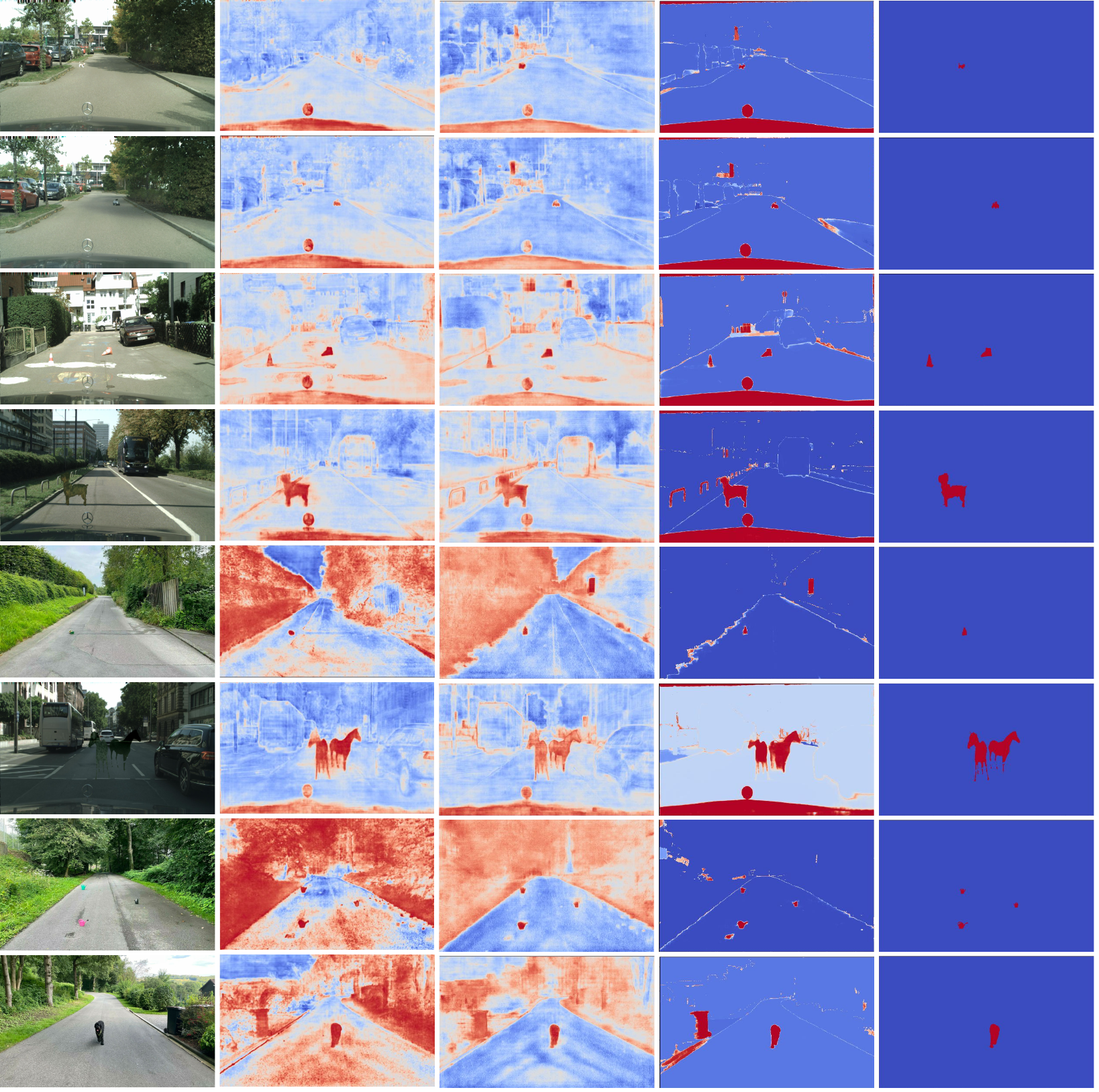}
    \end{center}
    \vspace{-1.25em}
    {{\hspace{3em}Input Image \hspace{2.15em}Maximized Entropy~\cite{chan2021entropy}\hspace{1.25em}Dense Hybrid~\cite{grcic2022densehybrid}\hspace{1.75em}Mask2Anomaly(Ours)\hspace{2.5em}Ground Truth}}
\caption{\textbf{Qualitative Results}: We observe that per-pixel classification architecture: Maximized Entropy and Dense Hybrid suffer from large false positives, whereas~\our{} which is a mask-transformer, show confident results across all datasets. Anomalies are represented in red.}
\label{fig:sup:f1} \vspace{-1em}
\end{figure*}

\begin{figure*}[t]
    \begin{center}
        \includegraphics[width=1\textwidth]{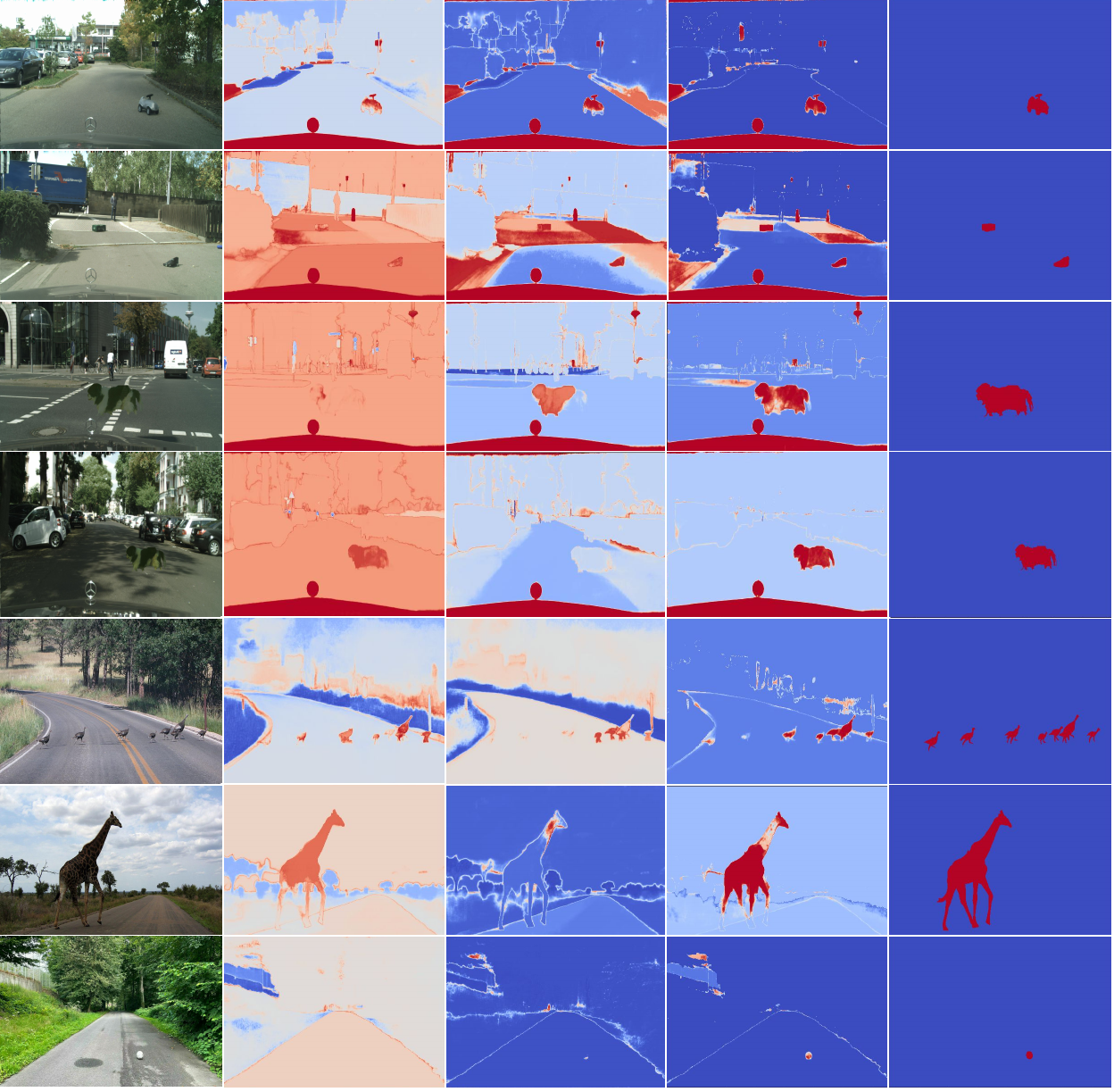}
    \end{center}
    \vspace{-1.25em}
    {\small{\hspace{3.25em}Input Image \hspace{4.25em}Cross-Attention~\cite{cheng2021per}\hspace{3.25em}Mask-Attention~\cite{cheng2022masked}\hspace{1.25em}Global Mask Attention (Ours)\hspace{2.75em}Ground Truth}}
\caption{\textbf{Attention Comparison}: We observe that the proposed global mask attention can better segment anomaly among the compared attention mechanism. Anomalies are represented in red.}
\label{fig:sup:f2} \vspace{-1em}
\end{figure*}

\begin{figure*}[t]
    \begin{center}
        \includegraphics[width=1\textwidth]{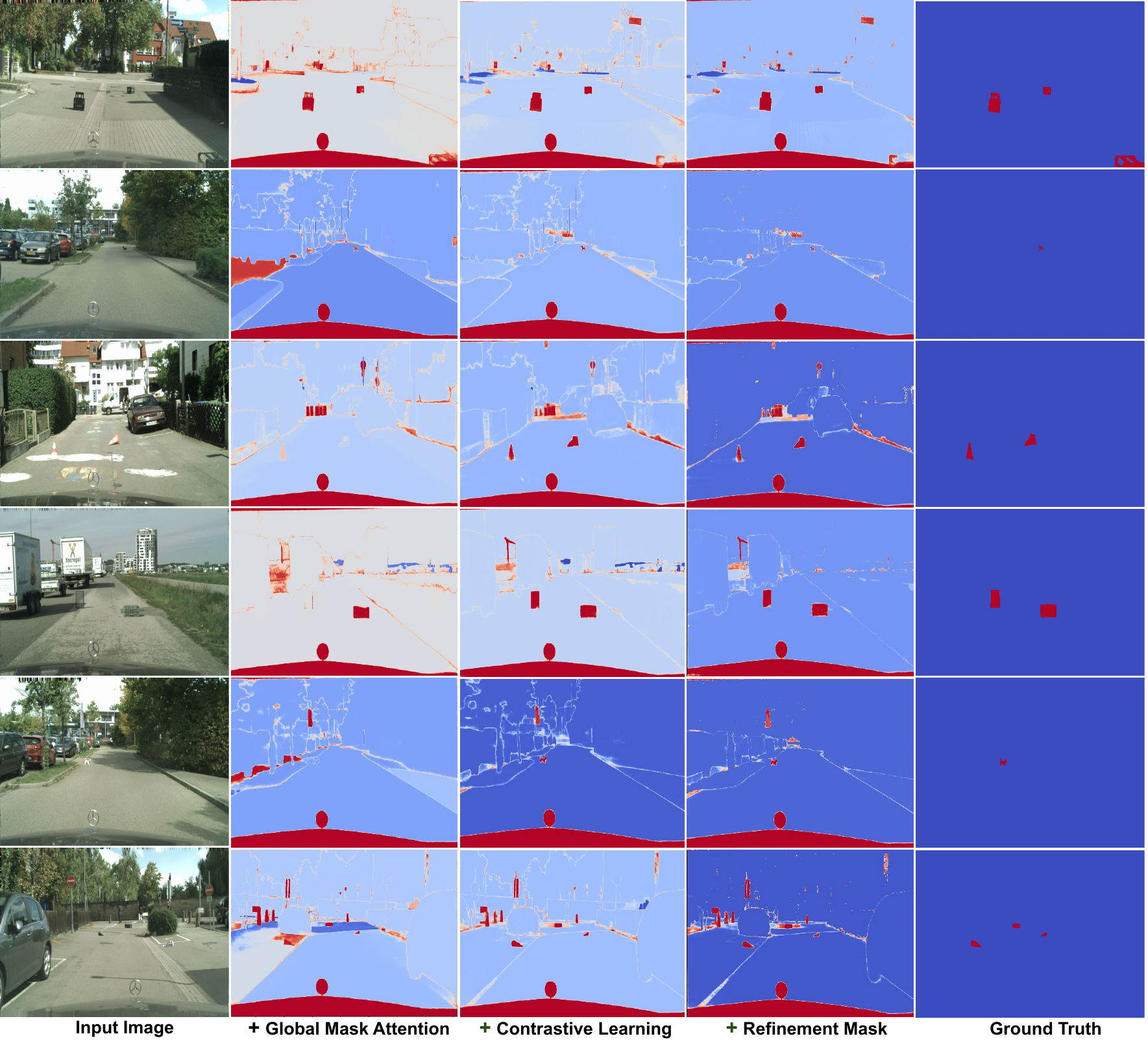}
    \end{center}
    \vspace{-1.25em}
\caption{\textbf{ Mask2Anomaly Qualitative Ablation}: shows the performance gain by progressively adding (left to right ) proposed components. Anomalies are represented in red.}
\label{fig:sup:f3} \vspace{-1em}
\end{figure*}

\begin{figure*}[t]
    \begin{center}
        \includegraphics[width=1\linewidth]{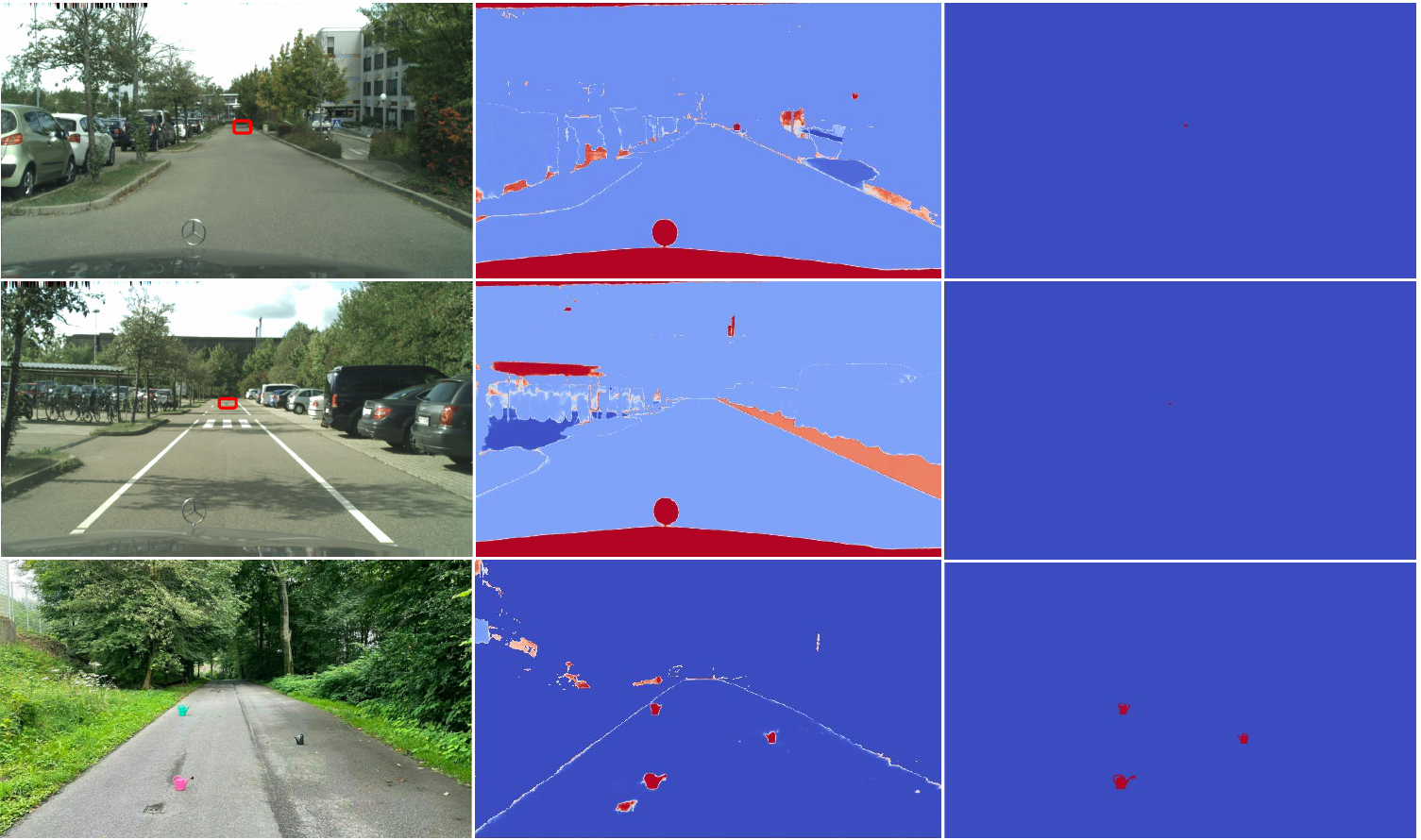}
    \end{center}
    \vspace{-1em}
    {\large{\hspace{4.5em}Input Image\hspace{9em}Mask2Anomaly\hspace{7em}Ground Truth}}
\caption{\textbf{Failure Cases:} Row (1,2): We can observe that Mask2Anomaly is unable to segment tiny anomalies(inside red bounding boxes of input image). Please zoom in for better clarity. Row 3: Mask2Anomaly falsely segments the pothole on the road as an anomaly. Anomalies are indicated in red in the ground truth.}
    \label{fig:failureCases}
    \vspace{-1em}
\end{figure*}

\myparagraph{G. Additional Results}\\
\textbf{Segmentation results:} In~\cref{sstab:main} and~\cref{fig:ss}, we show the segmentation results for~\our{} and Mask2Former. We can qualitatively and qualitatively infer that~\our{} performs better than Mask2Former.\\
\textbf{Qualitative anomaly segmentation:} In~\cref{fig:sup:f1}, we show the qualitative comparison of~\our{} with best-existing anomaly segmentation methods: Maximized Entropy~\cite{chan2021entropy} and Dense Hybrid~\cite{grcic2022densehybrid}. We observe that these per-pixel classification architectures suffer from large false positives, whereas~\our{}, a mask-transformer, shows confident results across all datasets. \\
\textbf{Attention comparison:}~\Cref{fig:sup:f2} shows the anomaly segmentation results obtained using various attention mechanisms, and the global mask attention clearly exhibits the best performance.\\ 
\textbf{Qualitative ablation study:} We show a component-wise qualitative ablation of~\our{} in ~\cref{fig:sup:f3} by progressively adding each components. We can observe that each proposed component improves anomaly segmentation and complements the others.\\
\textbf{Supplementary video:} Shows the performance of~\our{} on the sequence of images of small obstacle dataset~\cite{singh2020lidar}.~\our{} displays an impressive performance in segmenting wildlife on the road and anomalies in low-light conditions.\\
\textbf{Failure cases:} ~\cref{fig:failureCases} shows that~\our{} struggles to segment tiny anomalies and falsely detects road potholes as anomalies.


\end{document}